\documentclass[lettersize,journal]{IEEEtran}
\usepackage{amsmath,amsfonts}
\usepackage{algorithmic}
\usepackage{algorithm}
\usepackage{array}
\usepackage[caption=false,font=normalsize,labelfont=sf,textfont=sf]{subfig}
\usepackage{textcomp}
\usepackage{stfloats}
\usepackage{url}
\usepackage{verbatim}
\usepackage{graphicx}
\usepackage{cite}
\usepackage[table]{xcolor}
\usepackage{tablefootnote}
\usepackage{hyperref}

\usepackage{pifont}

\definecolor{lightgray}{RGB}{255, 255, 255}
\definecolor{darkgray}{RGB}{240, 240, 240}

\newcommand{\cmark}{\textcolor{black}{\ding{51}}}

\begin{document}

\title{Neural Radiance Fields in the Industrial and Robotics Domain: Applications, Research Opportunities and Use Cases}

\author{Eugen \v{S}lapak, Enric Pardo, Mat\'{u}\v{s} Dopiriak, Taras Maksymyuk and Juraj Gazda \thanks{Eugen \v{S}lapak, Mat\'{u}\v{s} Dopiriak and Juraj Gazda were with the Department of Computers and Informatics, Technical University of Kosice, Slovakia, e-mail:\{eugen.slapak, matus.dopiriak, juraj.gazda\}@tuke.sk.}
 \thanks{Enric Pardo was with Luxembourg Institute of Science and Technology, Luxembourg, e-mail: enric.pardo-grino@list.lu.}

\thanks{Taras Maksymyuk was with Department of Telecommunications, Lviv Polytechnic National University, Ukraine, e-mail: taras.maksymiuk@lpnu.ua.}
}

\markboth{Journal of \LaTeX\ Class Files,~Vol.~14, No.~8, August~2021}%
{Shell \MakeLowercase{\textit{et al.}}: A Sample Article Using IEEEtran.cls for IEEE Journals}

\maketitle

\begin{abstract}
The proliferation of technologies, such as extended reality (XR), has increased the demand for high-quality three-dimensional (3D) graphical representations. Industrial 3D applications encompass computer-aided design (CAD), finite element analysis (FEA), scanning, and robotics. However, current methods employed for industrial 3D representations suffer from high implementation costs and reliance on manual human input for accurate 3D modeling. To address these challenges, neural radiance fields (NeRFs) have emerged as a promising approach for learning 3D scene representations based on provided training 2D images. Despite a growing interest in NeRFs, their potential applications in various industrial subdomains are still unexplored. In this paper, we deliver a comprehensive examination of NeRF industrial applications while also providing direction for future research endeavors. We also present a series of proof-of-concept experiments that demonstrate the potential of NeRFs in the industrial domain. These experiments include NeRF-based video compression techniques and using NeRFs for 3D motion estimation in the context of collision avoidance. In the video compression experiment, our results show compression savings up to 48\% and 74\% for resolutions of 1920x1080 and 300x168, respectively. The motion estimation experiment used a 3D animation of a robotic arm to train Dynamic-NeRF (D-NeRF) and achieved an average peak signal-to-noise ratio (PSNR) of disparity map with the value of 23 dB and an structural similarity index measure (SSIM) 0.97.
\end{abstract}


\section{Introduction}






Early research on artificial intelligence (AI) applied to computer vision and 3D scene comprehension in robotics has primarily focused on simple algorithms, e.g., edge recognition for navigation \cite{kuipers2017shakey}. By the late 1990s, most algorithms lacked feature extraction capabilities from visual data. The growth in image feature extraction capabilities in robotics began with the appearance of the scale-invariant feature transform (SIFT) \cite{lowe1999object} and speeded up robust features (SURF) \cite{bay2006surf} approaches.

The recent development in machine learning has accelerated the progress of deep learning applications in robotics. One notable breakthrough was the successful implementation of deep learning based on the AlexNet convolutional neural network (CNN) \cite{krizhevsky2017imagenet}, which outperforms traditional computer vision methods in terms of performance. Since then, deep learning has been widely adopted in various robotic scenarios, enabling the integration of computer vision tasks with end-to-end trained robot control agents.

A comprehensive survey conducted in \cite{karoly2020deep} found that deep learning has been successfully applied in diverse areas of robotics, including grasp planning, path planning, sensory integration, and a broad range of deep reinforcement learning approaches. These applications have empowered robots to learn complex tasks, including joint planning and control models, as exemplified in works such as \cite{levine2016end}.

Furthermore, with the recent growth in the computational capacity of CPUs, GPUs and TPUs, we observe more industrial activities on digital twins and industrial metaverse applications. One prominent example is BMW's digital factory, where comprehensive 3D scanning has been utilized to accurately represent all production sites and digital twins that enable real-time, virtual navigation irrespective of geographical or temporal constraints \cite{prashar2023}. This technique allows the recreation of various operational elements, such as production lines, machinery and personnel. Furthermore, the developed virtual factory enables immersive engagement using virtual reality headsets to capture, analyze, and experience the workspace in the virtual domain. In general, using digital twins designed either by 3D scanning or by 3D modeling results in substantial savings in operational expenses across the whole industrial process.

The authors in \cite{vodrahalli20173d} provided a detailed review of 3D computer vision techniques in robotics based on deep learning, covering the period prior to 2017. The review focuses on two primary 3D computer vision task categories: 3D classification and generation. Classification tasks include the presence of an object class and estimating its 3D bounding box. Generation tasks involve transforming inputs into 2D or 3D representations of the 3D space. Examples of 3D representations include depth maps \cite{eigen2014depth} and voxel grids \cite{hane2017hierarchical, wu20153d}. The review also provided an overview of various neural architectures, primarily CNNs, which are trained using 2D image depth map channels or 3D representations such as voxels. These deep learning-based approaches perform better than previous methods for extracting depth and 3D representations from 2D images.

However, existing approaches have certain limitations, such as the need for manual human input for 3D modeling, which incurs high overhead costs and implementation time. Alternative approaches based on point clouds of scanned objects have extremely high storage and computational requirements. These drawbacks are driving further research to improve existing techniques.

Neural implicit representations, formally described in \cite{sitzmann2020implicit}, are a recent and notable paradigm in deep learning. Unlike traditional models that acquire generalized knowledge, neural implicit representations are designed to overfit training samples. This enables them to reconstruct samples with low error during inference, making them capable of storing multimedia data such as images, audio, and video within the neural network weights.

In this work, we focus on NeRF, an example of an implicit neural representation. NeRF learns a 3D scene representation by mapping 3D coordinates to their corresponding radiance and optical density based on sample images depicting the scene. We explore several industrial applications where NeRFs have been experimentally employed to solve various problems. Furthermore, we provide suggestions and insights regarding the future potential of NeRFs to address existing industrial challenges.

To date, several review articles have been published regarding NeRFs. Notable contributions include Gao et al. \cite{gao2022nerf}, Zhu et al. \cite{tewari2022advances}, and Tewari et al. [80]. However, it is worth noting that these reviews primarily concentrate on NeRFs and their derivatives, omitting the discussion of neural rendering techniques that utilize image-space 2D inference. Furthermore, an additional work \cite{xie2022neural} provided a comprehensive discussion on the broader subject of neural fields while delving into exploring NeRF.

Our work has several distinctive advantages that fill the gaps in the existing review papers in the field. First, it specifically focuses on applying NeRFs in industrial and robotics domains, highlighting the significant potential for expanding NeRF applications within this specific domain. This targeted approach allows for a deeper exploration of the unique challenges and opportunities present in industrial and robotics settings. Second, our work covers not only established NeRF applications but also introduces promising and innovative ideas that have the potential to drive further advanced research. Unlike existing reviews that may only touch upon these ideas briefly, our work explores them more extensively, shedding light on their potential implications and benefits. 

Our work goes beyond theoretical discussions by including proof-of-concept experiments. These experiments validate the feasibility and potential of the suggested ideas for further research. Including such experimental evidence strengthens our arguments regarding the applicability and effectiveness of the presented ideas. 

To conclude, this work closes two main research gaps. The first is the lack of review and application exploration papers for NeRFs in the industrial domain. The second is the shortcomings of traditional applied 3D graphics and representation approaches in this domain, which existing and novel NeRF-based approaches can solve.
 Our main contributions can be summarized as follows:
 \begin{itemize}
  \item   We comprehensively review existing approaches for computer graphics and 3D representations in industrial applications and outline their weaknesses.
  \item  We investigate the feasibility of NeRF approaches in industrial applications compared with the weaknesses of predominantly used nonNeRF approaches.
  \item  We conduct experimental evaluations of the novel NeRF applications studied in this work.
\end{itemize}

The structure of the remainder of this paper is outlined as follows. Section II explains the fundamental principles underlying basic NeRFs, specifically operating with static scenes. Section III discusses the classification of the main families of derivative NeRF variants and their relevance for areas of interest. Section IV offers an overview of the current state of 3D rendering and 3D scene representation applications in the industrial domain and the potential applicability of NeRFs in these applications. Section V presents a mathematical description of NeRFs and the evaluation metrics required for our proof-of-concept experiments. Sections VI and VII present our experiments with NeRF-based video compression and depth estimation for obstacle avoidance, respectively. Finally, the paper concludes in Section VIII, summarizing the main findings and contributions.

\section{Neural radiance fields (NeRFs): Fundamentals and workflow}

NeRFs \cite{mildenhall2021nerf} belong to the family of implicit neural representation models. Such models generally represent a single continuous entity, e.g., a signal describing a single image or properties of a single 3D scene. They can also be considered compression and a signal function approximation technique for such entities. NeRFs use fully connected deep neural networks to integrate a 3D static scene representation into the neural network weights. The purpose is to parametrize a signal in the form of images and overfit the data. The original NeRF implementation uses a basic multilayer perceptron (MLP) architecture-based model. This is a departure from most tasks for which overfitting is an unwanted phenomenon, as neural networks are usually used for generalization over a set of entities instead of overfitting over a single learned entity. The input is a set of images depicting a single 3D scene and its corresponding camera poses. The differentiable rendering function calculates the radiance, which is used to obtain the resulting pixels of a 2D image render. Optimizing neural scene representation is then guided by minimizing between the ground truth and generated image. NeRFs accomplish a task widely known as a novel view synthesis determined for generating 2D images from a specific pose or producing a 3D scene model. \par

\begin{figure*}[!h]
	\centering
	\includegraphics[width=0.9\textwidth]{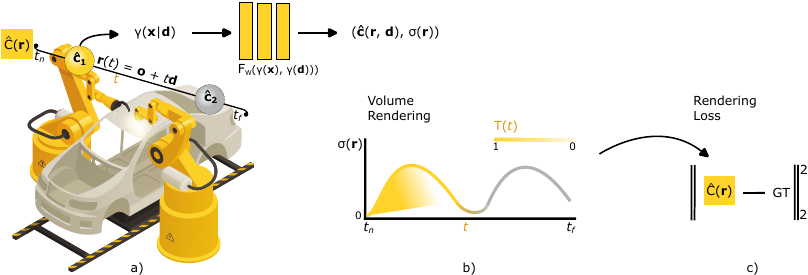}
    	\caption{(a) A 5D coordinate (spatial location in 3D coupled with directional polar angles) is transformed into a higher-dimensional space by positional encoding $\gamma$. It serves as an input for MLP $F_{W}$. The output from $F_{W}$ consists of gradually learned color $\hat{\boldsymbol{c}}$ and volume density $\sigma$ for the corresponding 5D input coordinate. $\boldsymbol{r}$, $\boldsymbol{d}$, $\boldsymbol{x}$ denote the ray vector, direction vector and spatial location vector, respectively. (b) Pixel values are obtained via volume rendering with numerically integrated rays bounded by respective near and far bounds $t_n$ and $t_f$. (c) Ground truth pixels from training set images are used to calculate rendering loss and optimize the $F_{W}$ weights via backpropagation \cite{mildenhall2021nerf}.
    }
 
	\label{fig:nerf_overview}
	
\end{figure*}

A visual overview accompanying the textual description of NeRF working principles, with some basic notations used throughout the paper, is available in Fig. \ref{fig:nerf_overview}. Compared to the majority of other methods, generating high-fidelity 3D static scenes or objects with NeRFs requires less effort with workflows similar to stereophotogrammetry \cite{do2019review}. The workflow involves training the NeRF using only an ensemble of input images that capture the desired 3D scene from different viewpoints. One limitation of the original NeRF architecture is that it needs additional information about the camera location and angle for each training image in the form of a camera transformation matrix. However, this shortcoming was successfully addressed by some of its derivative variants.

During the training process, NeRF enhances the accuracy and realism of its output by approximating the volume density and radiance of points in 3D space. These two quantities are crucial in defining object properties within a given scene. The volume density determines the opacity level of the matter at a specific point in space. Radiance determines the amount of radiant power emitted or reflected by an infinitesimally small projected area in watts per infinitesimal solid angle. These concepts are visually represented in Fig. \ref{fig:nerf_overview}a. It is necessary to consider radiance separately for individual red, green and blue color channels for color rendering.

While cameras capture 2D scene projections, each image may look quite different. NeRF learns one underlying 3D scene representation independent of any view transformations, which can be used to generate novel views unseen in the training set. Multiple training images allow the NeRF to learn a scene representation consistent for all training camera viewpoints. 

Images are synthesized from NeRF using traditional volumetric rendering, as shown in Fig. \ref{fig:nerf_overview}b. Volumetric rendering renders pixels of novel view images as a radiance numerically integrated along the ray paths that travel through points in a 3D scene, gradually accumulating radiance and ultimately entering the camera aperture. NeRF is queried only for volume density and radiance for relevant ray paths, which depend on the camera viewpoint, during the rendering process. 

Rendering results synthesized for camera viewpoints in the training set are compared to ground truth images to obtain a loss signal for gradient descent, as shown in Fig. \ref{fig:nerf_overview}c. This allows us to gradually optimize the MLP weights to better represent the scene.

\section{Classification of Derivative NeRF Implementations}

\begin{table*}[ht]
\centering
\caption{NeRF subtype classification (works covering multiple categories).}
\label{tab:nerf_classification}
\begin{tabular}{|p{7.5cm}|>{\columncolor{darkgray}}c|>{\columncolor{lightgray}}c|>{\columncolor{darkgray}}c|>{\columncolor{lightgray}}c|>{\columncolor{darkgray}}c|>{\columncolor{lightgray}}c|>{\columncolor{darkgray}}c|>{\columncolor{lightgray}}c|}
\hline
  &  \rotatebox{280}{Rendering speed eff.} & \rotatebox{280}{Training speed eff.} & \rotatebox{280}{Sample efficient} & \rotatebox{280}{Memory efficient} & \rotatebox{280}{Quality-focused} & \rotatebox{280}{Dynamic} \\ \hline
 
 \textbf{Neural sparse voxel fields} \cite{liu2020neural}, \textbf{KiloNeRF} \cite{reiser2021kilonerf}, \textbf{Adaptive Voronoi NeRFs} \cite{elsner2023adaptive}, \textbf{SNeRG} \cite{hedman2021baking} & \cmark & & & & & \\
 \textbf{Plenoctrees for real-time rendering of neural radiance fields} \cite{yu2021plenoctrees}, \textbf{FastNeRF} \cite{garbin2021fastnerf}, \textbf{Streaming radiance fields for 3d video synthesis} \cite{li2022streaming} &  & & & & & \\ \hline
 \textbf{Mega-NeRF} \cite{turki2022mega} & \cmark & \cmark & & & & \\ \hline
 \textbf{SqueezeNeRF} \cite{wadhwani2022squeezenerf}, \textbf{Mip-NeRF} \cite{barron2021mip} & \cmark & & & \cmark & & \\ \hline
 \textbf{Plenoxels} \cite{fridovich2022plenoxels} & & \cmark & & & &  \\ \hline
 \textbf{Depth-supervised NeRF} \cite{deng2022depth} & & \cmark & & & \cmark &  \\ \hline
 \textbf{TensoRF} \cite{chen2022tensorf} & & \cmark & & \cmark & \cmark &  \\ \hline
 \textbf{Fast dynamic radiance fields with time-aware neural voxels} \cite{fang2022fast}, \textbf{Devrf} \cite{liu2022devrf} & & \cmark & & & & \cmark  \\ \hline
  \textbf{Neural 3D Video Synthesis from Multiview Video} \cite{li2022neural}, \textbf{D-TensoRF} \cite{jang2022dtensorf} & & \cmark & & \cmark & & \cmark  \\ \hline
  \textbf{Temporal Interpolation Is All You Need for Dynamic Neural Radiance Fields} \cite{park2023temporal} & & \cmark & &  & \cmark & \cmark \\ \hline
 \textbf{RegNeRF} \cite{niemeyer2022regnerf}, \textbf{PixelNeRF} \cite{yu2021pixelnerf}, \textbf{LOLNeRF} \cite{rebain2022lolnerf}, \textbf{NeRDi} \cite{deng2022nerdi} \textbf{Semantically consistent few-shot view synthesis} \cite{jain2021putting} & & & \cmark & & & \\ \hline
 \textbf{CodeNeRF} \cite{jang2021codenerf} & & & \cmark & & & \cmark \\ \hline
  \textbf{HyperNeRF} \cite{park2021hypernerf}, \textbf{NeRF-DS} \cite{yan2023nerf} & & & & & \cmark & \cmark  \\ \hline
  \textbf{D-NeRF} \cite{pumarola2021d}, \textbf{Nerfies} \cite{park2021nerfies} & & & & & & \cmark \\ \hline
 
\end{tabular}
\end{table*}

This section summarizes derivative NeRF variants that enhance specific aspects of the original NeRF architecture in terms of efficiency or qualitative improvements. Several extensions have been developed to enhance various basic NeRF architecture properties, including training time, quality, required samples, and rendering time. Acquiring knowledge about these NeRF improvements is crucial to addressing the specific application constraints and requirements inherent to diverse scientific and practical contexts.

Each improvement is introduced by derivative NeRF variants addressing various constraints in industrial NeRF use cases.
For example, \emph{sample efficiency} can be useful if the robot needs to use NeRF for spatial navigation when moving to an unknown environment until enough images within a significant part of the environment from different viewpoints are collected. Low rendering times (\emph{performance oriented}) are useful for systems requiring low reaction times, for example, robots moving through space at high speeds. Finally, as \emph{dynamic} NeRF variants may be used to capture movement, one example of their utilization is learning of predefined repetitive movement of a robot for spatiotemporal coordination.

We categorize NeRF variants and structure the following subsections based on their main improvement focus over the original NeRF. These general categories are NeRF variants focused on \emph{sample efficiency}, \emph{performance} and \emph{system dynamics}. Although this classification helps to quickly determine the best NeRF variants for particular task, most papers focus on multiple improvements simultaneously. To reflect this, Table \ref{tab:nerf_classification} provides a more fine-grained classification of these works.

\subsection{Latent vectors for sample efficiency and scene modification}
Some NeRF variants that achieve sample efficiency leverage latent vectors that encode abstract features of object categories learned from diverse training data. Although latent vectors are not the only method for enhancing sample efficiency, as demonstrated by RegNeRF \cite{niemeyer2022regnerf}, they also enable scene modification in some NeRF architectures, such as CodeNeRF \cite{jang2021codenerf}.

\textbf{RegNeRF} \cite{niemeyer2022regnerf} enhances scene reconstruction from as few as three input images. Thus, it regularizes both the geometry and the appearance of the rendered image patches for novel views. Geometry regularization exploits the prior assumption of 3D geometry smoothness, which holds for most real-world scenes with low spatial frequency geometry. Appearance regularization optimizes the NeRF based on a novel view image patch, using its likelihood of correctness. Such likelihood is obtained using a normalizing flow model trained on an unstructured image dataset.

\textbf{pixelNeRF} \cite{yu2021pixelnerf} leverages local features extracted by a CNN to learn the correspondence between 2D and 3D features during training. The approach also employs a clustering algorithm to divide the image pixels into NeRF submodules, enabling parallel training.

\textbf{LOLNeRF} \cite{rebain2022lolnerf} performs 3D structure reconstruction from a single image of a given object class using generative latent optimization (GLO) with NeRF. The decoder network is trained with a latent code representation extracted from each image, which encodes abstract features relevant for 3D scene reconstruction. The latent code extraction is progressively refined during training.

\textbf{CodeNeRF} \cite{jang2021codenerf} learns disentangled latent codes for object 3D structure and texture during the training. It uses inference-time optimization over a single frame to adjust the latent code, which is used to condition the NeRF output. This enables inference-time modification of the object shape or texture to match a single image example without changing the MLP weights. Moreover, unlike similar approaches, knowledge of the camera pose is not required to perform this task.

\textbf{NeRFs from sparse RGB-D images for high-quality view synthesis} \cite{yuan2022neural} show that it can be trained from as few as 6 input images by pretraining and fine-tuning for each scene.

\textbf{ActiveNeRF} \cite{pan2022activenerf}: presents uncertainty-aware active learning for NeRFs and improves scene representation quality by choosing specific samples that reduce the NeRF uncertainty the most during training. In this way, it can use resources more efficiently by focusing on the most important samples.

\textbf{NeRDi} \cite{deng2022nerdi}: (Single-view NeRF synthesis with language-guided diffusion as general image priors) leverages multiview priors learned by 2D diffusion models to reconstruct 3D scenes from a single image. It uses image captioning and textual inversion from the input image to guide the coarse and fine object appearance, respectively.

\textbf{SparseNeRF} \cite{wang2023sparsenerf} uses imperfect depth information from low-cost depth sensors or depth estimation models to train NeRF with a small number of samples. It outperforms other state-of-the-art methods on the local light field fusion (LLFF) \cite{mildenhall2019local} and database transaction unit (DTU) \cite{jensen2014large} datasets.

\subsection{Performance}
Existing research works improve various NeRF performance metrics, such as rendering speed, training speed, and model memory footprint. Some methods achieve improvements on multiple metrics simultaneously. Efficiency enhancements also enable better performance and new practical applications on resource-constrained hardware, reducing deployment costs and enabling smaller hardware form factors useful for robotics.

\textbf{NSVF: Neural sparse voxel fields} \cite{liu2020neural} boosts rendering performance by 10x by training a set of implicit fields within voxels arranged in a voxel octree. The voxel structure is learned during training, and rendering is accelerated by skipping irrelevant voxels.

\textbf{KiloNeRF} \cite{reiser2021kilonerf} is a rendering performance-oriented method that splits the 3D scene volume into many smaller subvolumes, each served by a separate small MLP, instead of training a single large MLP on the whole 3D scene. This results in a rendering speedup of three orders of magnitude compared to the original NeRF.

\textbf{Adaptive Voronoi NeRF} \cite{elsner2023adaptive} uses a similar approach to KiloNeRF of dividing a single NeRF function into smaller functions. However, it uses Voronoi diagrams that evenly partition the geometry to be learned by a set of approximators into individual Voronoi cells. This method does not require a separate neural network distillation step like KiloNeRF and achieves better quality with 256 Voronoi cells than KiloNeRF with 512 regularly spaced cells.

\textbf{Mega-NeRF} \cite{turki2022mega} is an architecture designed for training on large 3D scenes such as unmanned aerial vehicle (UAV) flythroughs that capture areas as large as whole city blocks. It uses a sparse neural network with parameters specialized to local parts of the scene. It can achieve a 40 $\times$ speedup over basic NeRF with a negligible quality loss when using fast rendering NeRF variants on top of the Mega-NeRF architecture.

\textbf{SNeRG} \cite{hedman2021baking} introduces a data structure called the sparse neural radiance grid (SNeRG) that can store precomputed data in a sparse grid of voxels. Each voxel contains optical density, diffuse color, and learned specular color feature vectors computed by NeRF, which is used only during training. At inference time, diffuse color and optical density are alpha-composited along the ray. Specular color feature vectors are alpha-composited separately and fed to an MLP that outputs the specular color component. The diffuse and specular component images are then combined into the final image. This method accelerates the NeRF rendering on commodity hardware to 30 FPS. It also achieves space efficiency, requiring only 90 MB of storage.

\textbf{PlenOctrees} \cite{yu2021plenoctrees}: first trains a NeRF variant called NeRF-SH (SH stands for \emph{spherical harmonic}), which uses closed-form spherical basis functions instead of color. NeRF-SH is then precomputed into an octree-based 3D structure called PlenOctree. Color can be obtained by summing the weighted spherical harmonic bases for a given direction defined by the two view angles. PlenOctrees allow direct optimization for further quality improvement and threefold training speedup. They also achieve a 3000-fold rendering speed increase over the original NeRF at 150 FPS.

\textbf{Plenoxels} \cite{fridovich2022plenoxels} is a follow-up work to PlenOctrees, which uses a different voxel structure to represent the scene. Instead of an octree, it uses a dense 3D array of pointers to voxels. Each voxel edge defines a spherical harmonic, and the plenoptic function within each voxel is obtained by trilinear interpolation of the spherical harmonics. This contrasts with PlenOctree, which used a single constant value for the voxel volume. Plenoxels can perform end-to-end optimization without using NeRF, achieving a training speed approximately 100 $\times$ faster than the original NeRF. Therefore, we classify Plenoxels as mainly a training performance-focused approach based on our NeRF classification scheme.

\textbf{FastNeRF} \cite{garbin2021fastnerf}: factorizes the original NeRF MLP into two MLPs: one that depends on the position and one that depends on the view. The position-dependent MLP takes the 3D position coordinates as input and outputs a deep radiance map with $D$ components. The view-dependent MLP takes the two view angles $\theta$ and $\phi$ as input and outputs the radiance map component weights.

This factorization enables efficient caching and querying of radiance fields, resulting in a speed increase of approximately 3000 times over the original NeRF, with an average frame rate of approximately 200 FPS.

\textbf{SqueezeNeRF} \cite{wadhwani2022squeezenerf} further improves FastNeRF by splitting the location-dependent MLP into three separate MLPs, one for each axes pair. This reduces the caching memory requirements by 60 times compared to FastNeRF.

\textbf{Depth-supervised NeRF} \cite{deng2022depth} uses a sparse point cloud extracted from multiview images as an inexpensive signal to guide NeRF in scene reconstruction. This improves the scene reconstruction quality and increases the training speed by 2$\times$ to 3$\times$.

\subsection{Dynamic scene reconstruction with NeRF}

NeRF can produce photorealistic static scene reconstructions from multiple views. It can also be extended to handle dynamic scenes with temporal variations. Dynamic NeRF models can be useful for industrial applications that require 3D capture of repetitive processes. Most existing dynamic NeRF models focus on capturing past motion rather than predicting future scene motion. Many of them rely on the concept of deformation fields. Some approaches mentioned in this section were primarily motivated by human motion capture, where obtaining multiple perfectly static views is challenging. \par
\textbf{D-NeRF} \cite{pumarola2021d} encodes the canonical appearance of a static 3D scene at a reference time instant (usually $t$ = 0) with an MLP that learns the function $\Psi_{\mathrm{x}}(\mathbf{x}, \mathbf{d}) \mapsto(\mathbf{c}, \sigma)$, mapping the 3D location and view direction to the color and optical density pair. For other time instants, it uses another MLP to learn a deformation function that maps each canonical point in the scene to its new location $\Psi_t(\mathbf{x}, t)$. This function outputs displacement vectors $\Delta \mathbf{x}$ that correspond to the movement of each scene point. Such an approach is more effective than a single time-conditioned MLP that learns the scene appearance for each time instant separately.

\textbf{TiNeuVox} \cite{fang2022fast}: (Fast dynamic radiance fields with time-aware neural voxels) employs a small deformation network and multidistance interpolation to achieve comparable or better quality than D-NeRF, but with much faster training time, typically less than 10 minutes, compared to tens of hours required by D-NeRF.

\textbf{Nerfies} \cite{park2021nerfies}: address the challenge of human capture with NeRF, which is difficult due to inevitable human motion (e.g., breathing, muscle twitching, etc.) even when they attempt to remain still during multiple selfie shots. This motion causes misalignment of both $\sigma$ and $\mathbf{c}$ in space when projecting rays from different views. 
Unlike D-NeRF, which uses time latent vectors for deformation network training, Nerfies uses time and view transform-based latent vectors, as there may be multiple samples from the same view but different timestamps. 

\textbf{HyperNeRF} \cite{park2021hypernerf} extends Nerfies to handle topology changes, which are not well-captured by deformation fields due to the introduction of discontinuities by a changing topology. HyperNeRF projects NeRF into a higher-dimensional space, where each sample represents a slice through this space. It learns a higher-dimensional function that interpolates between slices. Inspired by the level set method, HyperNeRF uses curved cutting surfaces instead of straight hyperplanes, resulting in simpler shapes described by the higher-dimensional function. HyperNeRF outperforms Nerfies on interpolation (by 4.1\%) and novel view synthesis (by 8.6\%). 

\textbf{NeRF-DS} \cite{yan2023nerf} addresses the limitations of approaches based on deformation fields such as Nerfies and HyperNeRF for reconstructing dynamic specular objects (e.g., a shiny teapot or glassware). It reformulates the NeRF function to depend on the surface position and orientation and adds a moving objects mask to guide the deformation field.

\textbf{ DeVRF} \cite{liu2022devrf}: (Fast deformable voxel radiance fields for dynamic scenes) approach achieves a 100 $\times$ speedup in training compared to other dynamic NeRF models, such as D-NeRF, Nerfies, HyperNeRF and neural scene flow fields (NSFF). It targets inward-facing scenes and uses a multicamera setup for scene capture.

\textbf{LFNs}: LFNs (light field networks) are another line of research that explores neural representations based on neural light fields rather than radiance fields. A dynamic LFN model, DyLiN\cite{yu2023dylin}, learns from HyperNeRF, a high-quality but slow teacher. DyLiN outperforms both HyperNeRF and TiNeuVox, the current state-of-the-art dynamic model, in terms of the PSNR visual quality metric. Moreover, DyLiN is much faster than both approaches, achieving an order of magnitude speedup. Other recent works that address the dynamic LFN problem are bandlimited radiance fields (BLiRF) \cite{ramasinghe2023blirf}, RefiNeRF \cite{khalid2023refinerf}, temporal interpolation is all you need \cite{park2023temporal}, D-Tensorf \cite{jang2022dtensorf}, and DyNeRF \cite{li2022neural}.

\subsection{Other}

\textbf{Mip-NeRF}: Mip-NeRF \cite{barron2021mip} proposes an antialiasing technique that samples 3D scenes with cones instead of rays for each pixel, as in the original NeRF. This improves rendering quality for different camera distances and view resolutions, with an average error reduction of 17\% on the original NeRF dataset. Furthermore, the model achieves a 60\% error reduction and a 22 $\times$ rendering speedup compared to brute-force NeRF supersampling. The model also reduces the model size by half compared to the original NeRF implementation.

 \textbf{ICARUS}: ICARUS \cite{rao2022icarus} is an energy-efficient NeRF inference and rendering accelerator hardware architecture tested on FPGA.
 The current design of ICARUS achieves a NeRF rendering speed of 45.75 s per frame, with a power consumption of only 282.8 mW. The design is not parallelized yet, which could further improve the frame generation time. 

\section{Current status of industry and advantages of NeRF alternatives}

In this section, we cover the main areas where 3D representations and visualizations are applied with commercially used methods for such applications and specific recent NeRF-based approaches applied to these areas. 

\subsection{Computer-aided engineering}
Computer-aided engineering (CAE) refers to using computer software and tools to assist in engineering and design processes. It encompasses various techniques and methodologies that leverage computer technologies to analyze and simulate the behavior of products and systems.

Computer-aided design (CAD), as one particular branch of CAE, uses computer software to create and modify geometric models of physical objects. The main modeling approach used is freeform surface modeling. Freeform surfaces typically use boundary representation (B-Rep) based on nonuniform rational B-spline (NURBS) \cite{stroud2006boundary}. The resulting model can serve as an input for product dimension analysis, automated creation of technical drawings, bill of materials generation and other tasks. 

CAD models can be repurposed for various applications, particularly when specific techniques are employed to facilitate their reuse \cite{heikkinen2018review}. One notable application is simulation through finite element analysis (FEA), which allows for the calculation and interactive visualization of properties and interactions within products or manufacturing equipment. These simulations can encompass factors such as part stresses, temperatures, failure modes, fluid dynamics, and electromagnetic interactions, and they often uncover deficiencies or shortcomings in the CAD model, necessitating its modification. Thus, CAD models can also be updated based on modified mesh models utilized in FEA simulations \cite{louhichi2015cad}. 

The broader field of CAE encompasses a range of similar approaches, including the finite volume method (FVM) and finite difference method (FDM), computer-aided manufacturing (CAM) and computational fluid dynamics (CFD). These techniques collectively enable engineers and designers to analyze and optimize their designs, accounting for various factors and interactions, leading to improved products and systems.

In addition to product design, manufacturing processes need to be designed and optimized with as few expensive deployment and redeployment iterations as possible. 3D computer visualizations allow quick and error-free visual assessment of prototyped manufacturing facilities.

Emerging generative NeRFs that can generate the desired 3D model from scratch can partially supplement or even fulfill the role of CAD tools in the future. Applications of generative design include product design, training data generation for other models, or environment design, allowing another trained model to generalize better. Generative approaches developed for NeRF have different degrees of complexity and control available over the generated results. This mirrors a similar landscape of varying control options of existing 2D screen space image generation approaches, such as generative adversarial networks (GANs), diffusion models and multimodal transformer models. Generative NeRFs guided purely by natural language can produce diverse and creative outputs. However, they offer limited control over the properties of the outputs, which makes them suitable for only decorative and entertainment-oriented designs.

The text and shape-guided latent-NeRF approach \cite{metzer2022latent} offers enhanced control over the generated results compared to approaches solely conditioned on natural language. In addition to textual descriptions, this method provides shape guidance or an exact 3D shape for automatic texture generation. 
 
 The latent-NeRF approach is diffusion-based, which differs from conventionally training NeRF models in red, green and blue (RGB) color space. Instead, the NeRF model is trained in a latent space. This choice eliminates the need to convert images into latent space during training, resulting in reduced computational overhead.

  DreamFusion \cite{poole2022dreamfusion} is another diffusion-based method that converts text into NeRF 3D scenes. Since there are no sufficiently large training datasets of 3D models paired with their textual descriptions, the authors use the existing Imagen \cite{saharia2022photorealistic} diffusion-based generative model that works with only 2D image space as a prior. The Imagen model takes a shaded image rendered by NeRF linearly combined with random noise and produces a denoised version, which guides NeRF optimization in the pixel space. NeRF is optimized using this method to achieve low loss for NeRF views from random angles.
  
  Magic3D \cite{lin2022magic3d} improves upon DreamFusion by addressing two main limitations: slow NeRF optimization and low-resolution image guidance that reduces 3D model quality. It uses a two-stage approach. First, it uses a low-resolution diffusion model as a prior for creating a coarse 3D representation with Instant-NGP NeRF. In the second stage, it extracts a polygonal mesh and its texture from NeRF and further optimizes them using a differentiable renderer and a high-resolution diffusion prior.

   Some recent works have also proposed promising methods that can enable NeRF representations for engineering simulations, such as \cite{le2023differentiable} and \cite{li20223d}.

\subsection{SCADA systems}
Graphics tools that model manufacturing processes are essential not only during the planning phase before production begins but also for ongoing supervision. In many industrial settings, continuous monitoring and control by human operators are necessary to ensure optimal operation. This monitoring and control are typically facilitated through supervisory control and data acquisition (SCADA) systems, which employ graphical interfaces that enable both system monitoring and control functionalities.
Graphics utilized for SCADA interfaces range from commonly used 2D interactive schematics to three-dimensional graphics. In certain cases, incorporating three-dimensional visuals becomes imperative for accurately comprehending the system's state, or at the very least, it aids in enhancing the operator's ability to control the system.

Despite the aforementioned advantages of 3D SCADA graphical interfaces, 2D interfaces offer the advantage of being easier to design than their 3D counterparts. This is because 2D interfaces do not necessitate creating complex 3D polygonal models, controllers, and animations.

With NeRFs, creating 3D scenes depicting the system under control can be significantly simplified. Since SCADA systems present interactive graphics that visually reflect controlled system state changes, NeRF variants with modifiable scenes should be used. For example, when the operator remotely switches a piece of industrial machinery on or off, its color, lighting or similar indicator should change.

Some NeRF variants enable controlled modifications to 3D scenes, and one notable example is CodeNeRF \cite{jang2021codenerf}. CodeNeRF allows for precise changes to the texture and shape of objects in the 3D scene.

CLIP-NeRF \cite{wang2022clip} facilitates the modification of NeRF scenes by conditioning them on either text or image inputs. It leverages preexisting embeddings from the contrastive language-image pretraining (CLIP) model, which learns a joint embedding space for language and images.

One potential application of CLIP-NeRF is the ability to swiftly introduce real-world changes into a NeRF scene using reference images. For example, a photograph of a prototype for a new product can provide more detailed information than text alone in accurately describing the novel object properties. By utilizing CLIP-NeRF, these reference images can serve as valuable input for modifying the NeRF scene to incorporate the desired changes.

Dynamic NeRFs provide an alternative method to represent various controlled machinery states. They achieve this by showing different segments or individual frozen time instances of a 3D object's movement sequence. For instance, consider a 3D representation of a windmill: using dynamic NeRFs, the blades can be depicted as either static when querying dynamic NeRF for a single time instant or in motion, mirroring real-world windmill states.

\subsection{Training simulations for workforce}
Advanced 3D simulators are used to train workers in industrial environments. These simulators range from simple keyboard and mouse control with a regular computer monitor to virtual reality (VR), augmented reality (AR) and physical controllers modeled after the physical controls for the target application that mirror real-world interactions.

The effectiveness of AR and VR training for industrial maintenance and assembly tasks was evaluated in \cite{gavish2015evaluating} and compared with real-world training. The study found that AR training for such tasks should be encouraged, but VR needs further evaluation. However, many tasks and situations are too expensive, dangerous, or rare to be trained in the real world. For example, operators in the chemical industry are trained \cite{fracaro2021towards} in a virtual environment for this reason. In such cases, VR training is very useful, as there is no effective alternative.

Combining precisely controlled traditionally rendered graphics and NeRF can be used not only for training machine learning models, as demonstrated in NeRF2Real \cite{byravan2022nerf2real} but also for training humans.

Human central vision is sharpest in comparison with lower acuity peripheral vision. This is due to the central area of the retina containing a region of a dense array of light cones in an area called the fovea. For NeRFs used in real time by humans in a VR setting, conventional rendering and radiance field representation methods waste resources by unnecessarily rendering all regions of images sharply, including those intended for peripheral parts of human vision. FoV-NeRF \cite{deng2022fov} solves this problem, where foveated NeRFs that render images with nonuniform quality adjusted to central vision with the highest and peripheral vision with the lowest quality are presented. Such an approach decreases both the cost and latency of VR for workforce training or immersive SCADA systems. 

\subsection{Training simulations for machine learning}
   
    \textbf{Transfer of skills acquired in simulation}: The NeRF2Real \cite{byravan2022nerf2real} approach showcases the use of NeRF for rendering in a simulated environment. The first step involved creating a NeRF representation of a real robot's environment, where obstacles were extracted for collision modeling and integration with synthetic objects, including a dynamic ball. Additionally, NeRF was utilized for the robot's vision input and combined with the synthetic dynamic ball object. Subsequently, a 20-degree-of-freedom (DoF) humanoid robot learned the task of pushing the ball through reinforcement learning within this environment.

   {\textbf{Capture of environment for training of other navigation algorithms}}:\label{subsec:navigation} In the context of navigation algorithms, trained agents obtain view rendering outputs from NeRFs instead of interacting with a real 3D environment or relying on conventional polygonal rendering of a 3D environment. This is achieved by providing NeRFs with information regarding agent location and view angle changes resulting from its simulated actions \cite{adamkiewicz2022vision}. This capability extends to even massive environments, such as entire cities, by dividing the environment representation into overlapping NeRF blocks (referred to as block-NeRFs) \cite{tancik2022block}. With NeRFs, it is also possible to modify scene properties, such as achieving arbitrary scene relighting. This greatly expands the variety and quantity of realistic 2D image samples generated for a given 3D scene. For instance, models conditioned by text descriptions enable arbitrary changes to the object materials within the scene, enabling cost-effective training environment modifications to enhance diversity \cite{haque2023instruct}.

\textbf{Extracting quantitative physical properties}: Physical parameter extraction of objects in the scene based on NeRF shows interesting results, as provided in \cite{li2023pac}. The authors demonstrate that dynamic object properties such as friction angles, viscosity and Young`s modulus can be obtained via NeRF application. Such information can be collected for later use in simulations, with robots serving purely as data-collecting agents or directly used for real-time robot predictive capabilities and decision-making processes. 

\subsection{Positioning and navigation}
Several approaches are available for industrial positioning and navigation \cite{brena2017evolution}. These include ultrawideband (UWB) positioning, global positioning systems (GPS), differential GPS, and simultaneous localization and mapping (SLAM). However, these approaches have drawbacks, such as limited precision due to signal time-of-flight calculation imprecisions and error accumulation. Additionally, the equipment cost and data storage/transmission requirements for point cloud data in the case of LIDAR-based SLAM can be high.

  NeRF-based approaches that allow SLAM include bundle-adjusting neural radiance fields (BARFs) \cite{lin2021barf}. BARF is capable of coarse-to-fine registration and simultaneous optimization of the representation learned by NeRF and camera poses of training samples (images). This allows NeRF training from imperfect or missing camera poses.

   SLAM is a technique that requires 3D mapping of the environment. Therefore, methods that can reconstruct 3D scenes, such as NeRF, are often useful for SLAM. One example of a neural 3D surface reconstruction method for SLAM is the study by Sun et al. \cite{sun2021neuralrecon}. They propose a monocular time-coherent 3D reconstruction technique that uses a time window to capture the temporal consistency of the scene instead of reconstructing and merging individual frames.
    
   NICE-SLAM \cite{zhu2022nice} uses separate coarse, mid, and fine hierarchical feature grids for geometry and another feature grid for appearance. An example of a SLAM approach inspired by NeRFs is NeRF-SLAM \cite{rosinol2022nerf}, which uses an Instant-NGP-based hierarchical volumetric NeRF map.

 In another work by Zhu et al. \cite{zhu2022latitude}, a localization approach called LATITUDE (global localization with truncated dynamic low-pass filter) is introduced for city-scale NeRF. This method involves a coarse-to-fine localization strategy. Initially, a regressor trained on sample NeRF-generated images is used to obtain an initial estimation point for optimization in location estimation. To mitigate the risk of becoming trapped in local minima, the truncated dynamic low-pass filter (TDLF) is introduced as a solution.


\subsection{3D scanning and tomography} 3D scanning is a versatile technology with various applications in different industries. Some examples are: analyzing the terrain of mining or manufacturing sites, monitoring and managing resources such as biomass \cite{singh2016big}, inspecting for problems and defects, and performing SLAM. 3D scanning can also enable high-value customization and compatibility with existing objects. For instance, it can facilitate faster reverse engineering of legacy or third-party components, tailor-made prosthetics, and accurate damage geometry measurements for custom repair of high-value products or equipment (e.g., a cargo ship). However, storing and processing LIDAR data remains a challenge. A typical LIDAR scan can contain millions of discrete points, and scanning over large areas generates datasets that can be considered big data \cite{deibe2018big}, \cite{kitchin2016makes}. For large-scale volume estimates, statistics obtained from a small part of the area of interest are often extrapolated to the whole area \cite{singh2016big}.

 Knowledge of 3D scene geometry has a wide range of applications, such as 3D scanning of objects ranging from small components to large buildings and enabling robots to interact with their environment (e.g., avoiding obstacles and grip planning). One method for estimating geometry is to directly query NeRF for volume density, which has been shown to correlate well with the presence of matter \cite{adamkiewicz2022vision}. Alternatively, obstacle meshes can be extracted using cube marching or methods based on signed distance fields, such as \cite{rakotosaona2023nerfmeshing}, which offers improved quality of the extracted geometry.
    
    NeRFs offer a significant improvement in depth estimation for some challenging cases, such as transparent objects \cite{ichnowski2021dex}. This demonstrates the advantage of neural networks in learning subtle visual cues needed to infer the object geometry compared to other techniques. Moreover, it is possible to obtain 3D scenes from blurry images, with NeRF removing the blurriness during the training process \cite{ma2022deblur}. This approach enables faster data acquisition for NeRF training without requiring special equipment that can capture sharp images in motion, allowing, for example, collecting images over large areas by rapidly moving UAVs at low altitudes. The geometry estimation problems discussed in this subsection are particularly common in the geospatial industry and addressed by NeRF research (e.g., \cite{mari2022sat}).

    Several NeRF-based methods are designed specifically for human tracking and pose estimation. Unlike most other objects in industrial environments, humans are highly mobile, and their movements are potentially unpredictable. These human mobility characteristics demand accurate and fast real-time tracking and ideally predictive methods. \cite{weng2022humannerf} presents a fine-grained tracking of human 3D poses from monocular video, including parts of geometry not visible in the video.
    
    One specific approach to robotics design is to create cobots (collaborative robots) that emphasize collaboration between humans and robots \cite{peshkin2001cobot}. Such collaborative robotics require more fine-grained tracking and prediction of human movement than robots that need to only track humans for avoidance or collaborating loosely. Recently, several works have focused on NeRF-based human capture and tracking, which can help to address the challenges of achieving precise human-robot interaction in collaborative robotics settings. For example, \cite{gao2022mps} introduces a NeRF-based method of rendering novel human poses and novel 3D scene views with a given human reconstructed from a sparse set of multiview images.

Tomography has various applications in industrial settings, such as flaw detection, failure analysis, metrology, assembly analysis and reverse engineering. Tomography is also used for novel tasks, such as UAV-aided scanning of the spatial distribution of gas plumes \cite{neumann2019aerial}. NeRFs applied to sinograph data obtained from X-ray scanners can achieve state-of-the-art 3D reconstruction results, as demonstrated in \cite{zang2021intratomo}, where scaling beyond scanner resolution known as superresolution (up to 8 times increase in resolution), reconstruction from a small number of samples and reconstruction from limited angle tomography with a range of 45 degrees were achieved.

\subsection{Other areas}

    \textbf{Dense object descriptors}: 
     The paper \cite{yen2022nerf} presents a method for creating dense object descriptors that can be used for point correspondence and keypoint detection. This method is particularly useful for 6-DoF pick and place operations of thin and reflective objects, which have been challenging in the past. The proposed approach significantly outperforms recent off-the-shelf solutions such as GLU-Net \cite{truong2020glu}, GOCor \cite{truong2020gocor}, and PDC-Net \cite{truong2021learning}.

    \textbf{Point-set registration}: 
Aligning visual data from different sources or times is often required in various industries, such as geospatial and medical fields. This alignment enables identifying and comparing the same points or regions on a 3D object or scene in different images. For instance,  multiple tomographic scans can be used to evaluate physiological changes over time. Point registration is one method for achieving this alignment. Goli et al. \cite{goli2022nerf2nerf} used surface fields extracted from NeRFs to implement pairwise point registration. These surface fields are invariant to camera viewpoint variations. The optimization process then finds rigid transformations that align the surface fields, ensuring the alignment of the corresponding visual data.

\textbf{Synthetic Perturbations for Augmenting Robot Trajectories via NeRF (SPARTN)} \ \cite{zhou2023nerf} is a robot manipulation training approach that uses NeRF to generate perturbations and corresponding corrective actions for expert demonstrations of eye-in-hand robot manipulation tasks. This improves the imitation learning performance by 2.8 times.

\textbf{Synthetic dataset generation for supervised learning}: An approach that uses synthetic NeRF-generated images to train vision models is presented in Neural-Sim \cite{ge2022neural}. Variations in synthetic images include pose, zoom and lighting. Precise control over data used for training is achieved, and images that contribute to vision model loss the most are purposely generated. Thus, high-quality vision models tuned for images from specific domains can be obtained without collecting many potentially redundant real-world samples with limited variation.   

\textbf{Learned semantic relationships}: NeRF using the approach proposed in \textbf{JacobiNeRF} \cite{xu2023} can learn underlying abstract semantic relationships between scene entities such as objects or even individual pixels. Such a NeRF then exhibits semantically linked resonance in its output when its weights are perturbed along the gradient of a specific point in the scene. For example, if the brightness of a single point on a wall is changed, the brightness of other points on a given wall also changes, with other points in the scene remaining unperturbed. One JacobiNeRF application is decreasing the annotation burden since annotations can propagate to semantically linked points. This approach can generally be used for entity selection and scene modification.

\section{Requirements for Proof-of-Concept Experiments}

We conducted quantitative experiments to test the feasibility of some of our novel NeRF applications in our domain of interest and to investigate the methods needed for implementation.

The first experiment examined the tradeoffs between data savings and quality for compression with different settings, such as resolution and compression quality presets. We used Instant-NGP \cite{muller2022instant}, a fast NeRF variant, for this experiment, as it enabled quick iterations through different experimental configurations.

The second experiment explored using dynamic NeRFs for predictive navigation. We chose D-NeRF \cite{pumarola2021d}, a dynamic NeRF variant that can reconstruct high-quality dynamic scenes, for this experiment. Its significant advantage is an existing open source implementation that facilitates the rapid and reproducible setup of dynamic scene experiments.

The code for our experiments is publicly available at a GitHub repository \footnote{\url{https://github.com/Maftej/iisnerf}}

\subsection{Mathematical representation}  

We present the mathematical model for Instant-NGP and dynamic NeRF, the two algorithms we used in our experiments. We also explain the axis-aligned bounding box (AABB) cuboids used in these NeRF variants, which speed up the rendering process by limiting the rendered volume.

\subsubsection{Main definitions}

In mathematics, a field is a set that defines the behavior of some basic algebraic operations. A neural field is a field that can be parameterized by a neural network.

NeRFs are used to model a 3D scene reconstruction from a set 2D images.
The reconstruction is a neural field, denoted as $\Phi : X \rightarrow Y$, that maps each $\mathbf{x}_\mathrm{recon} \in X$ to corresponding field quantities $\mathbf{y}_\mathrm{recon}\in Y$. A camera sensor image observation can also be represented as a field $\Omega: S \rightarrow T$ that maps the sensor coordinates $S$, $\mathbf{x}_\mathrm{sens} \in S$ to the measurements $T, \mathbf{t}_\mathrm{sens}$ $\in T$. The forward map is a differentiable mapping between the two functions $F : (X \rightarrow Y ) \rightarrow  (S \rightarrow T)$.

Therefore, if the forward map is differentiable, we can solve the following optimization problem \cite{DBLP:journals/corr/abs-2111-11426} to find the neural field $\Phi$:

\begin{equation}
\mathrm{argmin}_{W} \int_{X \times S} || F( \Phi(\mathbf{x}_\mathrm{recon})) - \Omega(\mathbf{x}_\mathrm{sens}) ||.
\end{equation}
$W$ denotes the parameters of $\Phi$.
This general definition of the neural field also applies to various nonNeRF representations, such as neural signed distance fields.

Narrowing to NeRF, from a mathematical point of view, NeRF-specific $\mathbf{x}_\mathrm{recon}$ is a 5-tuple of $ (x,y,z,\theta,\phi)$ of 3D cartesian $x,y,z$ coordinates of a point sampled along the ray and view angles $\theta$, $\phi$ in a spherical coordinate system, which determine the direction of the ray from the camera. NeRF-specific $\mathbf{y}_\mathrm{recon}$ is a 4-tuple $(r,g,b,\sigma)$, where $r,g$ and $b$ are RGB radiance color components of the specific point sampled along the ray with a particular direction and $\sigma$ is an optical density at this given point. NeRF is thus a function that transforms a 5-D $\mathbf{x}_\mathrm{recon}$ vector into another 4-D $\mathbf{y}_\mathrm{recon}$ vector. This mapping can be expressed as:

\begin{equation}
\label{eq:nerf_to_radiance}
   F(W):(x,y,z,\theta,\phi) \implies (r,g,b,\sigma),
\end{equation}

In the volumetric rendering process, rays are emitted through each pixel to compute the resulting pixel color. Only ray segments that fall within the AABB bounding volume are used for rendering computations. In formal terms, the expected color $C(\textbf{r})$ of a camera ray $r(t) = \textbf{o} + t\textbf{d}$ defined by its origin $\textbf{o}$ and direction $\textbf{d}$ at a point $t$ along the ray, within near and far ray bounds $t_n$ and $t_f$, is defined as:

\begin{equation} \label{eq:expected_colour}
C(\textbf{r}) =\int_{t_n}^{t_f}T(t)\sigma(\textbf{r}(t))\textbf{c}(\textbf{r}(t),\textbf{d})dt, 
\end{equation}

\noindent, where $\textbf{c}$ is the emitted color at a ray angle determined by $\textbf{d}$ and at a particular point on the ray line segment, parametrized as parameter $t$. Function $T(t)$ is the transmittance, which determines the contribution of individual $\textbf{c}$ values to the total $C(\textbf{r})$. 

\subsubsection{Instant Neural Graphics Primitives}

One main concern of rendering a full image with NeRF is the computation time. Since the  NeRF methodology approach from the original NeRF paper, different attempts to speed up the process have been carried out. Instant neural graphics primitives (Instant NGPs) \cite{M_ller_2022} attempt to reduce the number of layers and exponentially decrease the number of computations in the MLP computations. The idea behind this is to consider 2 parameters: the number of parameters $T$ and the objective of the resolution we want to obtain $N_\mathrm{max}$. The so-called multiresolution hash encoding attempts to not only train the weights as usual MLP but also the training parameters. Each level is independent and feature vectors at the grid vertices, the resolution of which is chosen to be a geometric progression between the worst and clearest resolutions [$N_\mathrm{min}$, $N_\mathrm{max}$]:

\begin{equation}
    N_l = N_\mathrm{min} bl,
\end{equation}
where

\begin{equation} b=\exp\left(\frac{\ln N_\mathrm{max}-\ln N_\mathrm{min}}{L-1}\right), \end{equation}

and $b$ depends on the number of levels and the attempted resolution.

\subsubsection{D-NeRF}

 Dynamic NeRF (D-NeRF) is one NeRF variant capable of 3D dynamic scene capture.
 Because of the additional temporal dimension, the mapping for the dynamic NeRF variants must be modified from Eq. \ref{eq:nerf_to_radiance} to the following form:

\begin{equation}
   F(W):(x,y,z,\theta,\phi,t) \implies (r,g,b,\sigma).
\end{equation}

  Although the model learning the direct mapping from 6D to 4D could be used for this task, the results in \cite{pumarola2021d} show that splitting the mapping into two separate functions reduces the computation time.
  Specifically, one MLP $\Phi_x$ serves as the reference static scene representation for a single time instant of the dynamic scene, usually for $t=0$. Another MLP, $\Phi_t$, then learns dynamic scene deformations of a canonical scene conditioned on time. Both MLPs thus learn the scene appearance over the time interval of interest.

\subsubsection{AABB representation}

In general, an AABB is a 2D rectangular bounding area or a 3D cuboid bounding volume aligned with coordinate axes. It is used to simplify calculations for simulation tasks such as collision detection or ray intersection. In the context of NeRFs, cuboid AABB is used to cut off the ray propagation at AABB boundaries.

 AABB can be defined by two parameters: a center point and a half-extent vector. The center point has 3D Cartesian coordinates ($X_0$,$Y_0$,$Z_0$) and specifies the location of the cuboid. The half-extent vector has components ($X$,$Y$,$Z$) and specifies how far the cuboid extends from the center point in each direction. The cuboid dimensions are thus $2X$ units in width, $2Y$ units in height, and $2Z$ units in depth. In summary, the extent vector and its center uniquely define AABB, and it can be used for rendering acceleration. 

\subsection{Evaluation metrics}
We consider well-established computer vision domain evaluation metrics: PSNR and SSIM. 

\textbf{PSNR} is the ratio of the maximum signal power to noise power measured in dB. Both the original image before noise corruption and the noisy image are needed for the PSNR calculation. To calculate the noise power component, the mean square error (MSE) of the noisy image is used. For a grayscale image with dimensions of $m$ x $n$ pixels, such MSE is calculated in the following fashion \cite{hore2010image}:

\begin{equation}
\textit{MSE}=\frac{1}{m n} \sum_{i=0}^{m-1} \sum_{j=0}^{n-1}[I(i, j)-K(i, j)]^2 ,
\end{equation}
where $I$ and $K$ are functions describing the original and noisy image, respectively, by mapping pixels indexed by $i$ and $j$ to signal values. For grayscale images, signal values determine the overall pixel brightness, and for color images, RGB color channel brightness values.
For color images in RGB color space, MSE is simply calculated across all three color channels. 

The PSNR is then calculated as follows:

\begin{equation}
\textit{PSNR}=10 \cdot \log _{10}\left(\frac{M A X_{I}^2}{M S E}\right), 
\end{equation}
where $MAX_{I}$ denotes the peak signal value in the original noiseless image.

It is usually expressed in dB, which can be computed as:
\begin{equation}
\textit{PSNR (dB)}= 20 log_{10} (\textit{MAX}_I) - 10 log_{10} (\textit{MSE}).
\end{equation}


\textbf{SSIM} is the measure of structural similarity between two images based on the luminance, contrast, and structure of the images. As opposed to PSNR, this metric accounts for specifics of human visual perception i and provides a quality evaluation function closer to that of an average human. It is calculated by comparing local patches (rectangular windows) of pixels in the images. The formula for its calculation is given in the following equation \cite{hore2010image}:

\begin{equation} 
\operatorname{SSIM}(x, y)=\frac{\left(2 \mu_x \mu_y+c_1\right)\left(2 \sigma_{x y}+c_2\right)}{\left(\mu_x^2+\mu_y^2+c_1\right)\left(\sigma_x^2+\sigma_y^2+c_2\right)}, 
\end{equation} where $x$ and $y$ are two image patches of size $N \times N$, $\mu_x$ and $\mu_y$ are the average pixel values of $x$ and $y$, $\sigma_x^2$ and $\sigma_y^2$ are the variances of $x$ and $y$, $\sigma_{xy}$ is the covariance of $x$ and $y$, and $c_1$ and $c_2$ are two constants to avoid instability when the denominator is close to zero. SSIM ranges from -1 to 1, where 1 indicates perfect similarity, and -1 indicates perfect dissimilarity.

\section{Proof-of-concept I.: UAV video compression using Instant-NGP}
 
Our first proposed novel NeRF application is a video compression technique based on NeRFs. This approach encodes only the difference between the NeRF-generated expected 2D representation of a scene with a given camera pose and actual frames captured in real time. Such enhanced video compression reduces the bit rate needed for transmission, as only the data about camera pose and information about the transformation between the real and generated frames would be sent through the network. We conducted an experiment in a virtual industrial 3D scene and measured the compression savings achievable by a UAV transmitting video stream with our compression approach in this scenario.  

\begin{figure*}[!ht]
	\centering
	\includegraphics[width=1\textwidth]{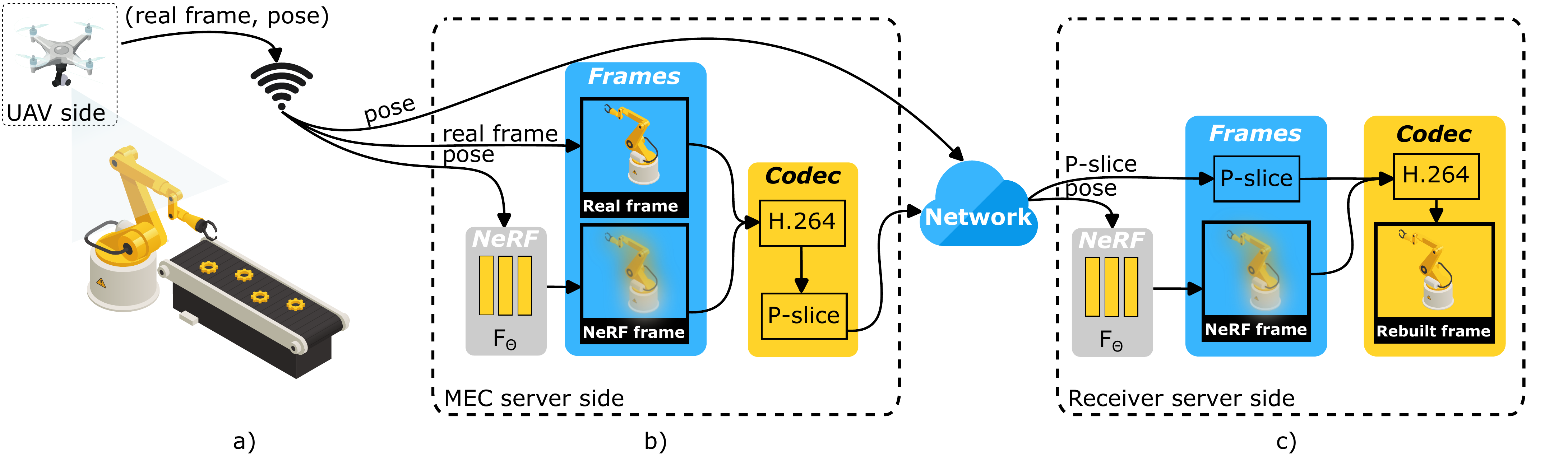}
	\caption{(a) A UAV camera captures the environment. The real frame and pose of the camera are transmitted wirelessly to a nearby multiaccess edge computing (MEC) server. (b) The MEC server employs the NeRF model for novel view synthesis based on camera pose. The H.264 codec encodes real and NeRF frames to obtain \emph{P frame} containing their differences, which is transferred through the network with the pose. (c) Receiver rebuilds the real frame using H.264 codec from \emph{P frame} and locally generated NeRF frame from camera pose.}
	\label{fig:nerf_video_compression}
\end{figure*}

\vspace{5mm}

\subsection{Proposed NeRF-based compression}

While widely used video compression algorithms use optical flow to encode the differences between frames, limitations increase the amount of information that needs to be added to each frame, as some frame changes cannot be described in terms of visual flow transformations of the previous frame \cite{fortun2015}. On the other hand, NeRFs can extract information about views obtained by a wide range of 3D transformations.

Due to NeRF`s knowledge of a full 3D scene, in contrast to conventional video compression, it has the knowledge of pixels unseen in all previously encoded frames. For example, the visual flow does not have access to information about unseen pixels beyond the edges of a view when view frustrum moves to show these in a video. Likewise, it does not have information about the occluded parts of objects when these become visible during object rotation. 
  
Importantly, the weights of a trained NeRF take less data storage space than a single 2D view image generated for a learned 3D scene \cite{mildenhall2021nerf}.

The H.264 compression algorithm was used in all our experiments. In H.264, the stream of compressed frames includes so-called \emph{I frames} and \emph{P frames}. \emph{I frames} consist of full image information, and \emph{P frames} code only the difference data between the image encoded in \emph{I frame} and its temporal neighbor image \cite{stockhammer2003h}. In compressed video, every $n$-th frame is an \emph{I frame} with $n$ conventionally determined as a constant value before compression starts.  

With NeRF pretrained on a given scene, the whole-frame information typically stored in the \emph{I frame} in compressed video can be omitted and generated by NeRF as needed from only camera pose information that takes less than 1 kB of data. For video streamed over the network, both the sender and receiver use a copy of the NeRF trained on the scene of interest, so both can use NeRF-generated images instead of sending the I frames. 

The suitable network architecture for UAV use-case scenarios involves computation load distribution in a MEC network so that NeRF-related computations can be offloaded from the UAV to the edge server.

The design of our compression approach and MEC network architecture is shown in Fig. \ref{fig:nerf_video_compression}. 

\subsection{Experimental design and results}

Primarily, we list all significant simulation parameters in Table \ref{tab:simulation_settings}. As a NeRF model, Instant-NGP \cite{muller2022instant} without any hyperparameter changes or architectural modifications was used. We prepared an experiment to demonstrate the performance of the novel NeRF-based compression algorithm in the industrial 3D scene depicted in Fig. \ref{fig:nerf_video_compression} and generated the training dataset for NeRF by rendering the industrial 3D scene in Blender 3D modeling software.
\begin{center}
\begin{table}[ht!]
\caption{Simulation parameters of proof-of-concept I.}
\label{tab:simulation_settings}
\begin{tabular}{ | m{4 cm} | m{4 cm}| } 

\hline
  NeRF variant &  Instant-NGP \cite{M_ller_2022}\\
\hline
  abbreviated Instant-NGP repository commit hash & 11496c211f\\
\hline
  AABB &  128\\
\hline
  3D modeling software &  Blender 3.4\\
\hline
  3D Renderer & Eevee\\
\hline
  Blender plugin for creating a synthetic dataset &  BlenderNeRF 3.0.0 \tablefootnote{https://github.com/maximeraafat/BlenderNeRF}\\
\hline
  BlenderNeRF dataset method & Subset of Frames (SOF)  \\
\hline
 third-party 3D models and textures & factory building \tablefootnote{https://www.turbosquid.com/3d-models/max-warehouse-pbr-gaming/1017253}, animated robotic exhaust pipe welding \tablefootnote{https://sketchfab.com/3d-models/demo-robotic-exhaust-pipe-welding-d3b8f5439e624ff18d26ca43ba302a0a}, garage doors in an original resolution \tablefootnote{https://www.pexels.com/photo/corrugated-metal-sheet-3312575/} \\
\hline
  scene lighting & HDR map in a 1K resolution \tablefootnote{https://polyhaven.com/a/derelict\_highway\_midday} \\
\hline
  training view count & 159 \\
\hline
  image resolutions used for training & $1920\times1080$ \\
\hline
  image resolutions used for evaluation of image quality and compression savings & $300 \times168$, $500\times280$, $720\times404$, $1920\times1080$ \\
\hline
  Constant Rate Factor (CRF) & 18, 23, 28 \\
\hline
  Preset & Veryslow, Medium, Veryfast \\
\hline
  
\end{tabular}
\end{table}
\end{center}

\begin{figure*}
	\centering
	\includegraphics[width=0.8\textwidth]{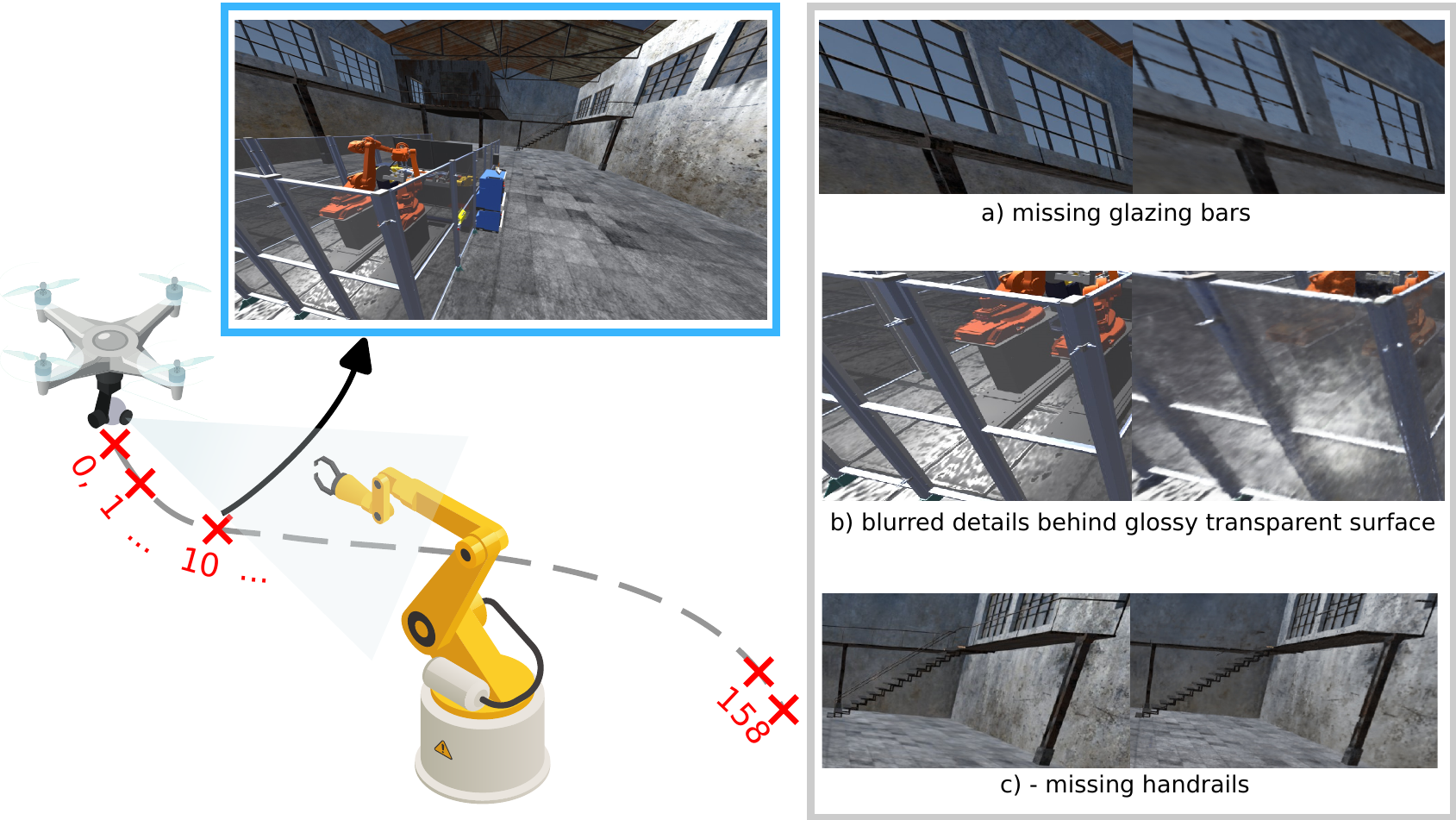}
	\caption{A UAV camera captures images at points illustrated on the dashed UAV trajectory. Images are indexed from 0 to 158, and this indexing is also used in Figs. 5-7. The frame with index 10 has the worst SSIM, as shown in Fig. \ref{fig:ssim}. Cropped areas from this frame, with their reconstruction by NeRF, are shown to highlight the causes of SSIM degradation.}
	\label{fig:drone_diagram}
\end{figure*}

An illustration of the experiment, with a UAV flying with a camera through the virtual environment, and an indexing scheme for frames captured along its trajectory are illustrated in Fig. \ref{fig:drone_diagram}. This figure also illustrates some quality degradation artifacts introduced during frame reconstruction by NeRF.   

\begin{figure}
	\centering
	\includegraphics[width=0.38\textwidth]{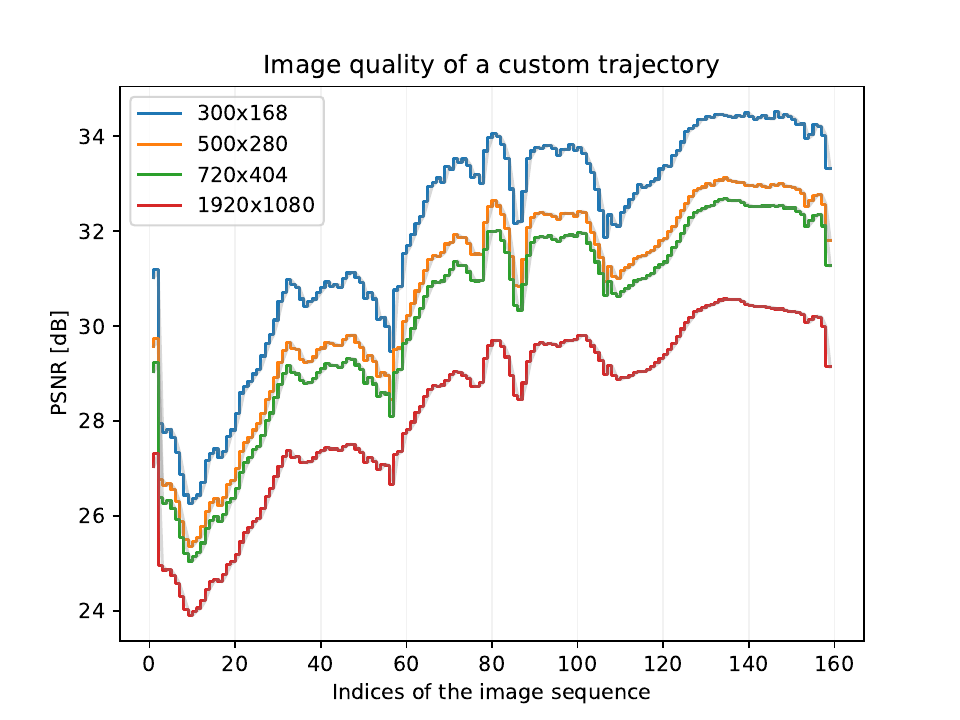}
	\caption{NeRF frame reconstruction quality along the UAV trajectory through the 3D scene measured with PSNR in dB.}
	\label{fig:psnr}
\end{figure}

Fig. \ref{fig:psnr} displays four measurement curves representing the PSNR metric, which evaluates the quality of the NeRF's reconstruction of 3D scene views compared to the original view quality. These measurements were obtained from individual frames captured during a simulated UAV trajectory, with each frame assigned an index ranging from 0 to 158. The PSNR curves correspond to the different resolutions listed in Table \ref{tab:simulation_settings}. Notably, the PSNR values vary across different indices, indicating varying reconstruction quality. Interestingly, as higher resolutions are attempted, the achievable PSNR decreases. However, it is worth mentioning that at resolutions of 500 and 720, the PSNR values are similar, showing a better tradeoff between PSNR and resolution in this resolution range.

\begin{figure}
	\centering
	\includegraphics[width=0.38\textwidth]{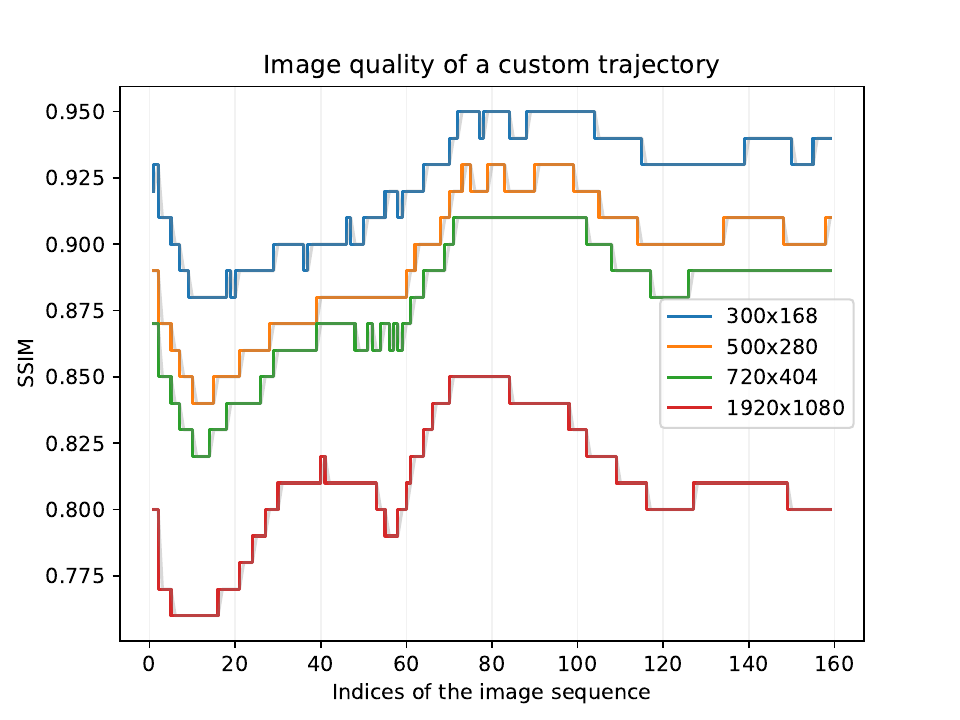}
	\caption{NeRF frame reconstruction quality along the UAV trajectory through the 3D scene measured with SSIM.}
	\label{fig:ssim}
\end{figure}

Fig. \ref{fig:ssim} illustrates the SSIM for the corresponding set of resolution objectives. The SSIM measurements exhibit different values; however, the overall behavior remains consistent, and all resolutions follow a similar trend to that of PSNR. Additionally, there are similarities in the SSIM achieved for the 500 and 720 resolutions, further supporting the notion that higher resolutions can be achieved without a substantial compromise in SSIM.

\begin{figure}[!htbp]
	\centering
	\includegraphics[width=0.38\textwidth]{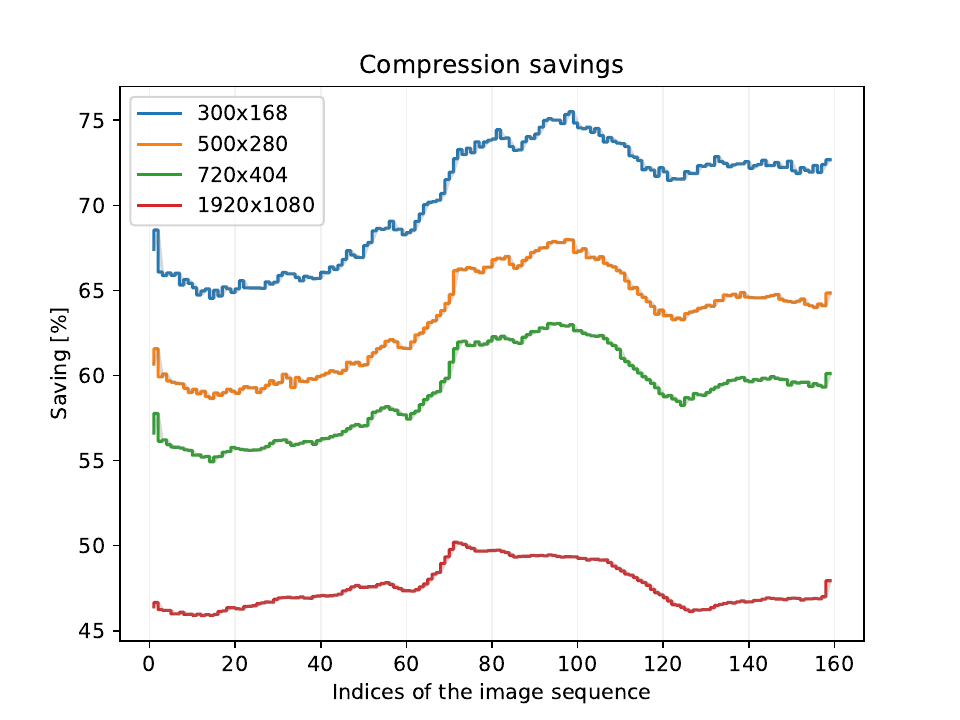}
	\caption{Compression savings relative to H.264 achieved for individual images along the UAV trajectory through the 3D scene. }
	\label{fig:compression_savings}
\end{figure}

Fig. \ref{fig:compression_savings} shows the percentage of compression savings for different resolutions. The compression savings of our approach are calculated relative to a set of frame pairs \emph{I fame} \emph{P frame} obtained by H.264 compression. We use the formula $ 100 * I_\mathrm{size} / (I_\mathrm{size} + P_\mathrm{size}) $ to calculate the compression savings, where $I_\mathrm{size}$ and $P_\mathrm{size}$ are the sizes of the \emph{I frame} and \emph{P frame}, respectively. Since we do not transmit \emph{I frame} over the network but generate it using NeRF, the percentage of $I_\mathrm{size}$ from the total of $I_\mathrm{size} + P_\mathrm{size}$ represents the savings.

We plotted the compression gains achieved by the proposed compression approach for 159 consecutive frames along the UAV trajectory for various frame resolutions from 300 $\times$ 168 to 1920 $\times$ 1080 (full HD). The main outcome is that the compression savings are inversely proportional to frame resolution. There are varying values of compression savings percentages from 45-48 \% for the resolution of 1920x1080 up to 66-74 \% when the resolution is lowered to 300x168.

The inverse proportionality of resolution and compression savings percentage is related to imperfections of NeRF-generated images. 
It has been observed that high-frequency signal components of both the original and NeRF-generated images are lost during downscaling. This phenomenon can be viewed as a form of lossy compression that brings the NeRF-generated image closer to the ground truth view by removing high-frequency artifacts introduced by NeRF. The behavior of SSIM supports these observations. It improves with decreasing resolution in a way that seems correlated with compression savings. For comparison, see corresponding SSIM and compression characteristics in Fig. \ref{fig:ssim} and Fig. \ref{fig:compression_savings}, respectively.

Relatively low values for the lower end of the resolution range we have used apply to multiple important scenarios. Should the video be used for visual machine learning downstream models, e.g., CNNs, the current state-of-the-art pretrained models typically work exclusively with relatively low resolutions, only up to 800 $\times$ 600, due to memory constraints.

\subsection{Computational and energy tradeoffs of NeRF-based video compression}

Our experiments show that using a NeRF can achieve significant compression savings but at the cost of higher computational and energy requirements than traditional compression techniques. This can be reduced by designing a suitable network architecture, such as MEC, that places computational resources close to communicating endpoints. 

While such a configuration requires UAVs to send a single UAV video stream without the benefit of our compression method to the MEC server, outgoing traffic from the server is still significantly lower. This offloads factory network nodes that aggregate the traffic from multiple links and typically experience the highest network load, for example, many such drone camera streams. It is generally less expensive and easier to provide high throughput and low latency locally at the network edges, where there are no issues with such traffic aggregation.

Depending on the computational capabilities of the sender and receiver, there are multiple possible computational offloading configurations. NeRF computations can be offloaded to an edge server close to the sender of the compressed video, receiver, or both. No offloading is needed if the sender or receiver is not constrained by battery or computational resources. However, since in our simulation scenario the sender is a battery-constrained UAV, it is beneficial to offload the NeRF-based compression workload to the edge server and send a high-throughput stream.

\section{Proof-of-concept II. : Obstacle avoidance using D-NeRF}

Regarding the dynamic NeRF proof-of-concept experiment, we train the D-NeRF on images of a 3D animation of the repetitive robotic arm movements. Due to the repetitiveness of these movements, D-NeRF can be used to provide predictions on future arm movement based on past arm locations. 

\begin{figure*}[!htbp]
	\centering
	\includegraphics[width=1\textwidth]{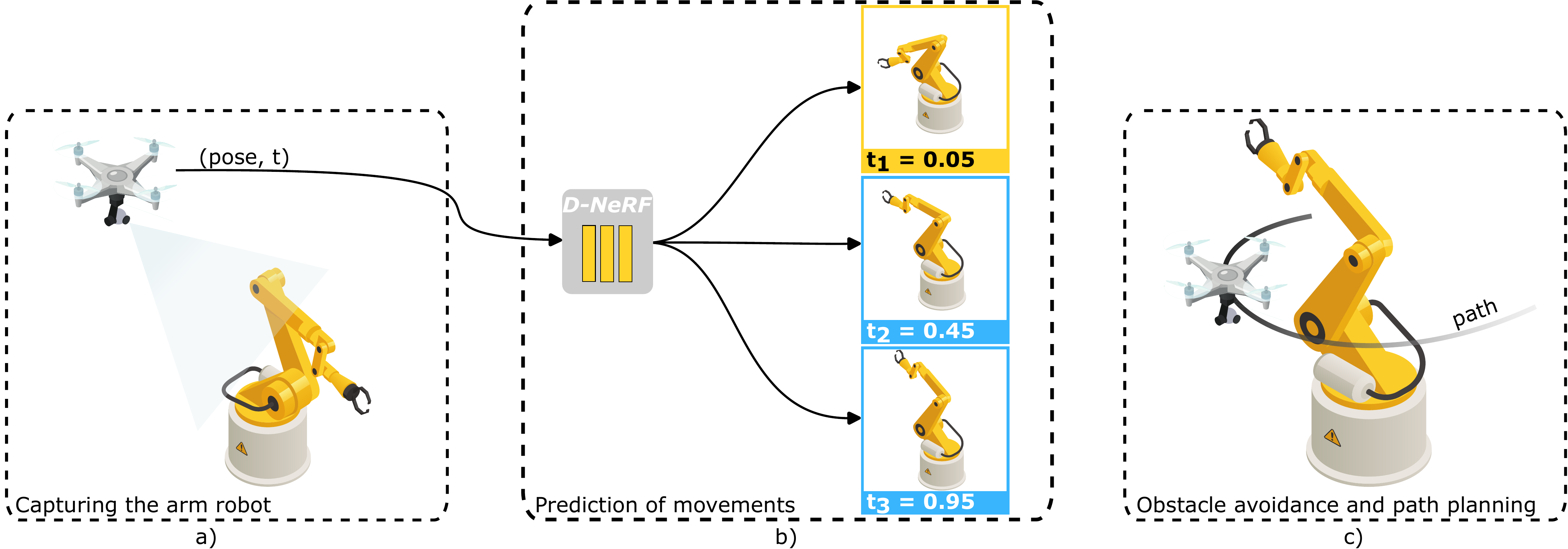}
	\caption{a) A UAV with a specific camera pose captures the object at time $t$. b) The D-NeRF model outputs novel views from the camera pose at specified time $t$ in the form of an RGB map, opacity map and disparity map depicted in Fig. \ref{fig:arm_robot_maps}. c) The opacity map and disparity map are key components for UAVs to perform obstacle avoidance or path planning.} 
	\label{fig:obstacle_avoidance_overview}
\end{figure*} 

\subsection{Proposed D-NeRF-based obstacle avoidance} 
There are multiple scene geometry extraction methods for individual time instants of dynamic scenes from D-NeRF that can be used for obstacle avoidance. Since approaches such as ray-marching are relatively slow, we use a disparity map for a set of rays. A disparity map is a measure of the apparent distance or motion of pixels between a pair of stereo images that can be used to estimate the depth of a scene. The disparity map thus helps us to estimate the distance in which the ray intersecting a given pixel intersects the surface of the closest obstacle. Since using this information to detect or predict collisions with UAV bounding volumes is trivial, we omit collision detection from our experiments and directly focus on the utility and quality of the generated disparity maps. 

The movement predictions need to be synchronized with the real-time arm 3D pose, matching it with one of the past poses on the learned dynamic scene timeline. Several methods that are beyond the scope of our work can be used for pose determination, e.g., \cite{heindl20193d}. We aim to show the possibility of avoiding collisions between the robotic arm and mobile robot (in our experiment, UAV). This use case can be obviously extended to other industrial processes with repetitive movement patterns. Our D-NeRF-based obstacle avoidance design is shown in Fig. \ref{fig:obstacle_avoidance_overview}.

\subsection{Experimental design and results}

Primarily, we list all significant simulation parameters in \ref{tab:dynamic_sim_settings}. For D-NeRF training, views of rendered robotic arm animation were collected with a virtual camera with a circular trajectory around the robotic arm, simulating the trajectory of the UAV. To obtain depth data describing the robotic arm surface usable for collision detection, a D-NeRF-generated disparity map was considered. The original disparity map generated from the trained D-NeRF contains a certain amount of visual noise visible as irregular spots. They appear both far and close to the robotic arm outline. There are several possible approaches to removing such noise \cite{fan2019brief}, \cite{ilesanmi2021methods}, \cite{bindal2022systematic}, and for this particular use case, we employed the image processing technique proposed in \cite{jamil2008noise}.
\begin{center}
\begin{table}[ht!]
\caption{Simulation parameters of proof-of-concept II.}
\label{tab:dynamic_sim_settings}
\begin{tabular}{ | m{4 cm} | m{4 cm}| } 
\hline
NeRF variant &  D-NeRF \cite{pumarola2021d}\\
\hline
  abbreviated D-NeRF repository commit hash &  89ed431fe1 \\
\hline
number of training iterations &  750 000\\
\hline
  3D modeling software &  Blender 3.4\\
\hline
  3D Renderer & Eevee\\
\hline
  plugin for creation of a disparity map dataset in Blender & Light Field Camera 0.0.1 \tablefootnote{https://github.com/gfxdisp/Blender-addon-light-field-camera}\\
\hline
 third-party 3D models &  animated robotic arm \tablefootnote{https://sketchfab.com/3d-models/black-honey-robotic-arm-c50671f2a8e74de2a2e687103fdc93ab}\\
  \hline
  training / testing / validation view count & 123 / 21 / 21 \\
\hline
image resolutions used for training images and RGB / opacity / disparity maps & 800x800 \\
  \hline
  
\end{tabular}
\end{table}
\end{center}
We illustrate trained D-NeRF renders, noise removal mask and obtained depth data at two time instants $t_1=0.05$ and $t_3=0.95$. 

\begin{figure*}[!htbp]
	\centering
	\includegraphics[width=1\textwidth]{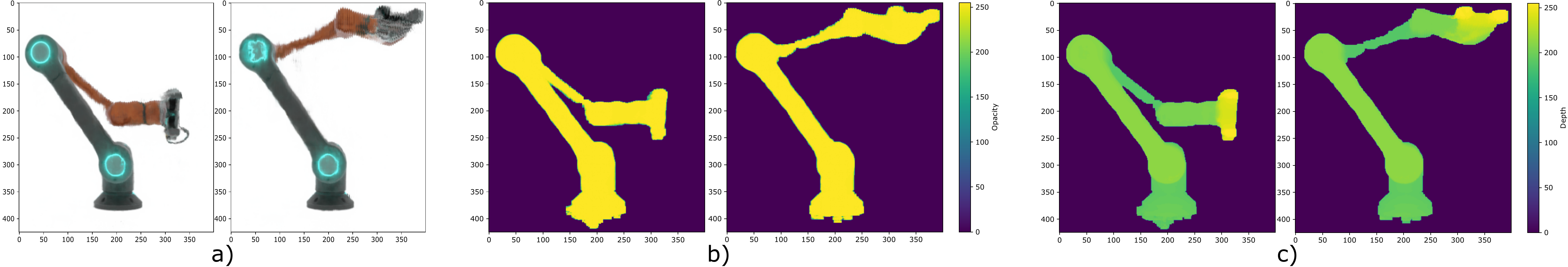}
	\caption{All maps extracted from the D-NeRF model representing the robot arm, rendered for two time instants $t_1=0.05$ and $t_3=0.95$. a) The RGB maps depict the novel view synthesis. b) Opacity maps show light ray transmissions through the environment, which can be transparent, partially or fully opaque. c) Processed disparity maps capture the robot arm surface depth during repetitive movements at these time instants as a heatmap.} 
\label{fig:arm_robot_maps}
\end{figure*}

Fig. \ref{fig:arm_robot_maps}a shows D-NeRF renders of these time instants. To obtain the final cleaned disparity map, visualized in Fig. \ref{fig:arm_robot_maps}c, we first generated an opacity map, which provided a higher level and quality of arm outline detail than the disparity map but also contained minor noise. Then, we applied a simple sequence of erosion and dilation OpenCV operations to the opacity map to create a clean opacity mask visualized in Fig. \ref{fig:arm_robot_maps}b, which is usable as a crop mask to crop the original noisy disparity map. Opacity map erosion removed most noise that was not part of the robotic arm outline, as the robotic arm is the largest continuous object in the scene. Most pixels eroded from the robotic arm outline were readded during the dilation step. We refer to the disparity map cropped by the denoised opacity map as the \emph{processed disparity map}. 

\begin{figure}[!h]
	\centering
	\includegraphics[width=0.38\textwidth]{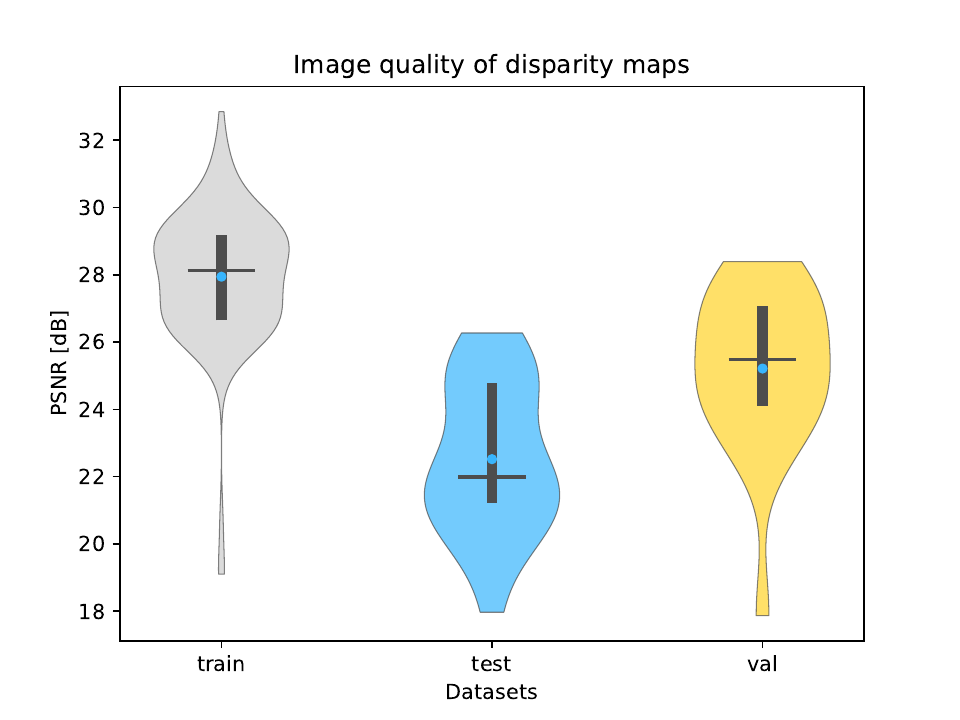}
	\caption{Comparison of disparity maps between Blender and D-NeRF images using the PSNR metric.}
	\label{fig:plot_dnerf_psnr}
\end{figure}

The quality of D-NeRF-generated disparity maps measured as PSNR relative to ground truth disparity maps for consecutive frames along the training, test and validation trajectories through the dynamic scene is plotted in Fig. \ref{fig:plot_dnerf_psnr}. To visualize the multimodal data distribution, kernel density estimation is employed, represented by gray, blue, and yellow colors for the training, testing and validation datasets, respectively. The wider sections within the visualization indicate a higher probability of the presence of certain values. The blue dot denotes the median, the horizontal line signifies the average mean, and the box plot depicts the interquartile range, providing insights into the spread of the middle half of the distribution. We can see that frames included in the D-NeRF training set deviate from the ground truth less for both measured quality metrics.

\begin{figure}[!h]
	\centering
	\includegraphics[width=0.38\textwidth]{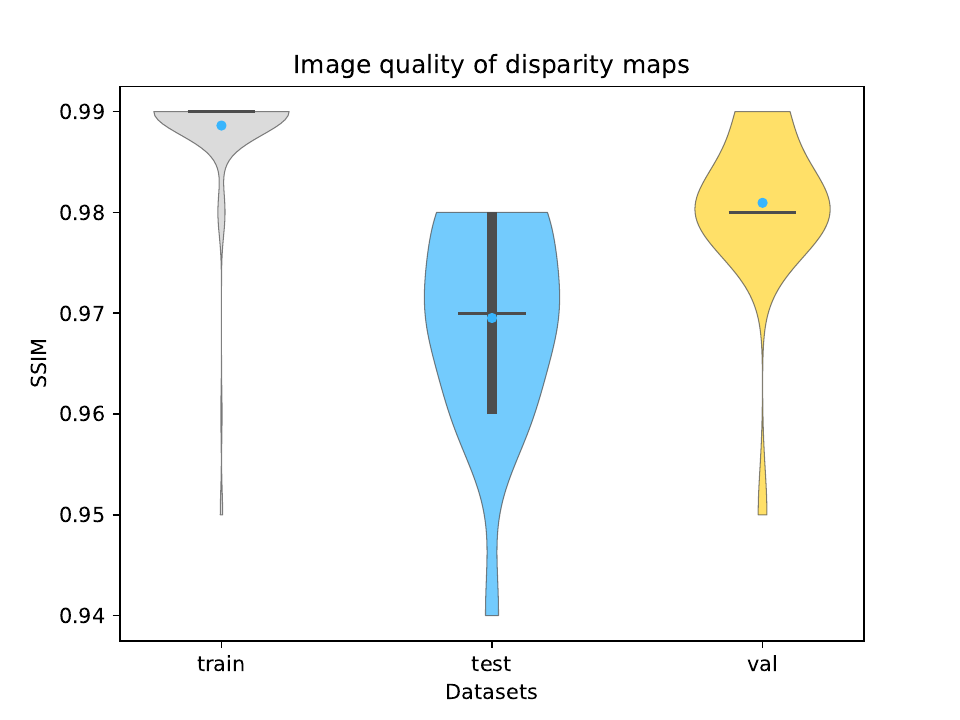}
	\caption{Comparison of disparity maps between Blender and D-NeRF images using the SSIM metric.}
	\label{fig:plot_dnerf_ssim}
\end{figure}

The results depicted in Fig. \ref{fig:plot_dnerf_ssim} demonstrate a significantly higher average quality of depth reconstruction, as measured by the SSIM, compared to the PSNR. For context, it is important to note which values are considered good for SSIM versus PSNR. This range for SSIM is from 0.97 to 1.0, while for PSNR, it is from 30 dB to 50 dB.

In the context of collision avoidance, the focus of the SSIM metric on the reconstruction quality of larger structures becomes crucial. This is due to collision bounding volumes, which only roughly approximate the geometry of obstacles and thus provide a tolerable margin of error against minor depth estimation deviations. The PSNR metric is more sensitive to such small-scale deviations; however, perfect precision is rarely needed for collision avoidance. Moreover, future research endeavors can explore how these disparities between PSNR and SSIM might influence other specific downstream tasks, thereby providing valuable insights for further investigations.

\section{Conclusion}
In this study, we conducted a comprehensive examination and explored the potential NeRF applications in the industrial sector. Our investigation has revealed that NeRFs can overcome limitations associated with traditional 3D representations and rendering methods, such as their high cost, storage inefficiency, equipment requirements, and limited realism. Furthermore, we investigated innovative NeRF applications in industrial settings and substantiated their viability through proof-of-concept experiments. We consider our work to be a valuable addition to the expanding body of literature concerning NeRFs and their diverse applications across various domains. Moreover, we aspire to facilitate bridging the gap between academic research and industrial practice by providing practical guidance and recommendations for employing NeRFs to address real-world industrial challenges.

As a novel and promising approach, NeRFs offer a wide range of opportunities for future research in the industrial domain. Our proof-of-concept experiments can be extended to explore the predictive speculative generation of video frames based on the prediction of future robot camera poses or even the predictive use of such frames for other downstream tasks, such as object classification. Speculative predictive execution could radically decrease the latency of such tasks. Beyond extensions of our experiments, there are still numerous unexplored topics. These include modeling the radio spectrum by NeRFs for signal quality estimation or generating views for individual robots in robot swarms from a video stream of static external cameras observing the swarm and its surroundings. Language-embedded NeRFs, such as the one presented in \cite{kerr2023lerf}, can also enable language model-based agents capable of commonsense reasoning for correct action planning in 3D.

\section*{Acknowledgment}
This work was supported by the Slovak Research and Development Agency under Grant APVV SK-CZ-RD-21-0028, the Ministry of Education, Science, Research and Sport of the Slovak Republic, and the Slovak Academy of Sciences under Grant VEGA 1/0685/23.

We would also like to acknowledge that Fig. 1, 2, 3 and 7, were made using licensed vector graphics available at Freepik.com, all used vector graphics was made by creator with username Macrovector.

\bibliographystyle{IEEEtran}
\bibliography{NeRF_Gazda_draft}

\begin{thebibliography}{100}
\providecommand{\url}[1]{#1}
\csname url@samestyle\endcsname
\providecommand{\newblock}{\relax}
\providecommand{\bibinfo}[2]{#2}
\providecommand{\BIBentrySTDinterwordspacing}{\spaceskip=0pt\relax}
\providecommand{\BIBentryALTinterwordstretchfactor}{4}
\providecommand{\BIBentryALTinterwordspacing}{\spaceskip=\fontdimen2\font plus
\BIBentryALTinterwordstretchfactor\fontdimen3\font minus
  \fontdimen4\font\relax}
\providecommand{\BIBforeignlanguage}[2]{{%
\expandafter\ifx\csname l@#1\endcsname\relax
\typeout{** WARNING: IEEEtran.bst: No hyphenation pattern has been}%
\typeout{** loaded for the language `#1'. Using the pattern for}%
\typeout{** the default language instead.}%
\else
\language=\csname l@#1\endcsname
\fi
#2}}
\providecommand{\BIBdecl}{\relax}
\BIBdecl

\bibitem{kuipers2017shakey}
B.~Kuipers, E.~A. Feigenbaum, P.~E. Hart, and N.~J. Nilsson, ``{Shakey: From
  conception to history},'' \emph{AI Magazine}, vol.~38, no.~1, pp. 88--103,
  2017.

\bibitem{lowe1999object}
D.~G. Lowe, ``{Object recognition from local scale-invariant features},'' in
  \emph{{Proceedings of the seventh IEEE international conference on computer
  vision}}, vol.~2.\hskip 1em plus 0.5em minus 0.4em\relax Ieee, 1999, pp.
  1150--1157.

\bibitem{bay2006surf}
H.~Bay, T.~Tuytelaars, and L.~Van~Gool, ``{Surf: Speeded up robust features},''
  \emph{Lecture notes in computer science}, vol. 3951, pp. 404--417, 2006.

\bibitem{krizhevsky2017imagenet}
A.~Krizhevsky, I.~Sutskever, and G.~E. Hinton, ``{Imagenet classification with
  deep convolutional neural networks},'' \emph{Communications of the ACM},
  vol.~60, no.~6, pp. 84--90, 2017.

\bibitem{karoly2020deep}
A.~I. K{\'a}roly, P.~Galambos, J.~Kuti, and I.~J. Rudas, ``{Deep learning in
  robotics: Survey on model structures and training strategies},'' \emph{IEEE
  Transactions on Systems, Man, and Cybernetics: Systems}, vol.~51, no.~1, pp.
  266--279, 2020.

\bibitem{levine2016end}
S.~Levine, C.~Finn, T.~Darrell, and P.~Abbeel, ``{End-to-end training of deep
  visuomotor policies},'' \emph{The Journal of Machine Learning Research},
  vol.~17, no.~1, pp. 1334--1373, 2016.

\bibitem{prashar2023}
A.~Prashar, G.~L. Tortorella, V.~R. Sreedharan \emph{et~al.}, ``Role of
  organizational learning on industry 4.0 awareness and adoption for business
  performance improvement,'' \emph{IEEE Transactions on Engineering
  Management}, 2023.

\bibitem{vodrahalli20173d}
K.~Vodrahalli and A.~K. Bhowmik, ``{3D computer vision based on machine
  learning with deep neural networks: A review},'' \emph{Journal of the Society
  for Information Display}, vol.~25, no.~11, pp. 676--694, 2017.

\bibitem{eigen2014depth}
D.~Eigen, C.~Puhrsch, and R.~Fergus, ``{Depth map prediction from a single
  image using a multi-scale deep network},'' \emph{Advances in neural
  information processing systems}, vol.~27, 2014.

\bibitem{hane2017hierarchical}
C.~H{\"a}ne, S.~Tulsiani, and J.~Malik, ``{Hierarchical surface prediction for
  3d object reconstruction},'' in \emph{{2017 International Conference on 3D
  Vision (3DV)}}.\hskip 1em plus 0.5em minus 0.4em\relax IEEE, 2017, pp.
  412--420.

\bibitem{wu20153d}
Z.~Wu, S.~Song, A.~Khosla, F.~Yu, L.~Zhang, X.~Tang, and J.~Xiao, ``{3d
  shapenets: A deep representation for volumetric shapes},'' in
  \emph{{Proceedings of the IEEE conference on computer vision and pattern
  recognition}}, 2015, pp. 1912--1920.

\bibitem{sitzmann2020implicit}
V.~Sitzmann, J.~Martel, A.~Bergman, D.~Lindell, and G.~Wetzstein, ``{Implicit
  neural representations with periodic activation functions},'' \emph{Advances
  in Neural Information Processing Systems}, vol.~33, pp. 7462--7473, 2020.

\bibitem{gao2022nerf}
K.~Gao, Y.~Gao, H.~He, D.~Lu, L.~Xu, and J.~Li, ``{NeRF: Neural radiance field
  in 3d vision, a comprehensive review},'' \emph{arXiv preprint
  arXiv:2210.00379}, 2022.

\bibitem{tewari2022advances}
A.~Tewari, J.~Thies, B.~Mildenhall, P.~Srinivasan, E.~Tretschk, W.~Yifan,
  C.~Lassner, V.~Sitzmann, R.~Martin-Brualla, S.~Lombardi \emph{et~al.},
  ``{Advances in neural rendering},'' in \emph{{Computer Graphics Forum}},
  vol.~41, no.~2.\hskip 1em plus 0.5em minus 0.4em\relax Wiley Online Library,
  2022, pp. 703--735.

\bibitem{xie2022neural}
Y.~Xie, T.~Takikawa, S.~Saito, O.~Litany, S.~Yan, N.~Khan, F.~Tombari,
  J.~Tompkin, V.~Sitzmann, and S.~Sridhar, ``{Neural fields in visual computing
  and beyond},'' in \emph{{Computer Graphics Forum}}, vol.~41, no.~2.\hskip 1em
  plus 0.5em minus 0.4em\relax Wiley Online Library, 2022, pp. 641--676.

\bibitem{mildenhall2021nerf}
B.~Mildenhall, P.~P. Srinivasan, M.~Tancik, J.~T. Barron, R.~Ramamoorthi, and
  R.~Ng, ``{NeRF: Representing scenes as neural radiance fields for view
  synthesis},'' \emph{Communications of the ACM}, vol.~65, no.~1, pp. 99--106,
  2021.

\bibitem{do2019review}
P.~N.~B. Do and Q.~C. Nguyen, ``{A review of stereo-photogrammetry method for
  3-D Reconstruction in Computer Vision},'' in \emph{{2019 19th International
  Symposium on Communications and Information Technologies (ISCIT)}}.\hskip 1em
  plus 0.5em minus 0.4em\relax IEEE, 2019, pp. 138--143.

\bibitem{liu2020neural}
L.~Liu, J.~Gu, K.~Zaw~Lin, T.-S. Chua, and C.~Theobalt, ``{Neural sparse voxel
  fields},'' \emph{Advances in Neural Information Processing Systems}, vol.~33,
  pp. 15\,651--15\,663, 2020.

\bibitem{reiser2021kilonerf}
C.~Reiser, S.~Peng, Y.~Liao, and A.~Geiger, ``{KiloNeRF: Speeding up neural
  radiance fields with thousands of tiny mlps},'' in \emph{{Proceedings of the
  IEEE/CVF International Conference on Computer Vision}}, 2021, pp.
  14\,335--14\,345.

\bibitem{elsner2023adaptive}
T.~Elsner, V.~Czech, J.~Berger, Z.~Selman, I.~Lim, and L.~Kobbelt, ``{Adaptive
  Voronoi NeRFs},'' \emph{arXiv preprint arXiv:2303.16001}, 2023.

\bibitem{hedman2021baking}
P.~Hedman, P.~P. Srinivasan, B.~Mildenhall, J.~T. Barron, and P.~Debevec,
  ``{Baking neural radiance fields for real-time view synthesis},'' in
  \emph{{Proceedings of the IEEE/CVF International Conference on Computer
  Vision}}, 2021, pp. 5875--5884.

\bibitem{yu2021plenoctrees}
A.~Yu, R.~Li, M.~Tancik, H.~Li, R.~Ng, and A.~Kanazawa, ``{Plenoctrees for
  real-time rendering of neural radiance fields},'' in \emph{{Proceedings of
  the IEEE/CVF International Conference on Computer Vision}}, 2021, pp.
  5752--5761.

\bibitem{garbin2021fastnerf}
S.~J. Garbin, M.~Kowalski, M.~Johnson, J.~Shotton, and J.~Valentin,
  ``{FastNeRF: High-fidelity neural rendering at 200fps},'' in
  \emph{{Proceedings of the IEEE/CVF International Conference on Computer
  Vision}}, 2021, pp. 14\,346--14\,355.

\bibitem{li2022streaming}
L.~Li, Z.~Shen, Z.~Wang, L.~Shen, and P.~Tan, ``Streaming radiance fields for
  3d video synthesis,'' \emph{Advances in Neural Information Processing
  Systems}, vol.~35, pp. 13\,485--13\,498, 2022.

\bibitem{turki2022mega}
H.~Turki, D.~Ramanan, and M.~Satyanarayanan, ``{Mega-NeRF: Scalable
  construction of large-scale nerfs for virtual fly-throughs},'' in
  \emph{{Proceedings of the IEEE/CVF Conference on Computer Vision and Pattern
  Recognition}}, 2022, pp. 12\,922--12\,931.

\bibitem{wadhwani2022squeezenerf}
K.~Wadhwani and T.~Kojima, ``{SqueezeNeRF: Further factorized FastNeRF for
  memory-efficient inference},'' in \emph{{Proceedings of the IEEE/CVF
  Conference on Computer Vision and Pattern Recognition}}, 2022, pp.
  2717--2725.

\bibitem{barron2021mip}
J.~T. Barron, B.~Mildenhall, M.~Tancik, P.~Hedman, R.~Martin-Brualla, and P.~P.
  Srinivasan, ``{Mip-NeRF: A multiscale representation for anti-aliasing neural
  radiance fields},'' in \emph{{Proceedings of the IEEE/CVF International
  Conference on Computer Vision}}, 2021, pp. 5855--5864.

\bibitem{fridovich2022plenoxels}
S.~Fridovich-Keil, A.~Yu, M.~Tancik, Q.~Chen, B.~Recht, and A.~Kanazawa,
  ``{Plenoxels: Radiance fields without neural networks},'' in
  \emph{{Proceedings of the IEEE/CVF Conference on Computer Vision and Pattern
  Recognition}}, 2022, pp. 5501--5510.

\bibitem{deng2022depth}
K.~Deng, A.~Liu, J.-Y. Zhu, and D.~Ramanan, ``{Depth-supervised NeRF: Fewer
  views and faster training for free},'' in \emph{{Proceedings of the IEEE/CVF
  Conference on Computer Vision and Pattern Recognition}}, 2022, pp.
  12\,882--12\,891.

\bibitem{chen2022tensorf}
A.~Chen, Z.~Xu, A.~Geiger, J.~Yu, and H.~Su, ``{Tensorf: Tensorial radiance
  fields},'' in \emph{{Computer Vision--ECCV 2022: 17th European Conference,
  Tel Aviv, Israel, October 23--27, 2022, Proceedings, Part XXXII}}.\hskip 1em
  plus 0.5em minus 0.4em\relax Springer, 2022, pp. 333--350.

\bibitem{fang2022fast}
J.~Fang, T.~Yi, X.~Wang, L.~Xie, X.~Zhang, W.~Liu, M.~Nie{\ss}ner, and Q.~Tian,
  ``{Fast dynamic radiance fields with time-aware neural voxels},'' in
  \emph{{SIGGRAPH Asia 2022 Conference Papers}}, 2022, pp. 1--9.

\bibitem{liu2022devrf}
J.-W. Liu, Y.-P. Cao, W.~Mao, W.~Zhang, D.~J. Zhang, J.~Keppo, Y.~Shan, X.~Qie,
  and M.~Z. Shou, ``{Devrf: Fast deformable voxel radiance fields for dynamic
  scenes},'' \emph{Advances in Neural Information Processing Systems}, vol.~35,
  pp. 36\,762--36\,775, 2022.

\bibitem{li2022neural}
T.~Li, M.~Slavcheva, M.~Zollhoefer, S.~Green, C.~Lassner, C.~Kim, T.~Schmidt,
  S.~Lovegrove, M.~Goesele, R.~Newcombe, and Z.~Lv, ``{Neural 3D Video
  Synthesis from Multi-view Video},'' 2022.

\bibitem{jang2022dtensorf}
H.~Jang and D.~Kim, ``{D-TensoRF: Tensorial Radiance Fields for Dynamic
  Scenes},'' 2022.

\bibitem{park2023temporal}
S.~Park, M.~Son, S.~Jang, Y.~C. Ahn, J.-Y. Kim, and N.~Kang, ``{Temporal
  Interpolation Is All You Need for Dynamic Neural Radiance Fields},'' 2023.

\bibitem{niemeyer2022regnerf}
M.~Niemeyer, J.~T. Barron, B.~Mildenhall, M.~S. Sajjadi, A.~Geiger, and
  N.~Radwan, ``{RegNeRF: Regularizing neural radiance fields for view synthesis
  from sparse inputs},'' in \emph{{Proceedings of the IEEE/CVF Conference on
  Computer Vision and Pattern Recognition}}, 2022, pp. 5480--5490.

\bibitem{yu2021pixelnerf}
A.~Yu, V.~Ye, M.~Tancik, and A.~Kanazawa, ``{pixelNeRF: Neural radiance fields
  from one or few images},'' in \emph{{Proceedings of the IEEE/CVF Conference
  on Computer Vision and Pattern Recognition}}, 2021, pp. 4578--4587.

\bibitem{rebain2022lolnerf}
D.~Rebain, M.~Matthews, K.~M. Yi, D.~Lagun, and A.~Tagliasacchi, ``{LolNeRF:
  Learn from one look},'' in \emph{{Proceedings of the IEEE/CVF Conference on
  Computer Vision and Pattern Recognition}}, 2022, pp. 1558--1567.

\bibitem{deng2022nerdi}
C.~Deng, C.~Jiang, C.~R. Qi, X.~Yan, Y.~Zhou, L.~Guibas, D.~Anguelov
  \emph{et~al.}, ``{NeRDi: Single-View NeRF Synthesis with Language-Guided
  Diffusion as General Image Priors},'' \emph{arXiv preprint arXiv:2212.03267},
  2022.

\bibitem{jain2021putting}
A.~Jain, M.~Tancik, and P.~Abbeel, ``{Putting NeRF on a diet: Semantically
  consistent few-shot view synthesis},'' in \emph{{Proceedings of the IEEE/CVF
  International Conference on Computer Vision}}, 2021, pp. 5885--5894.

\bibitem{jang2021codenerf}
W.~Jang and L.~Agapito, ``{CodeNeRF: Disentangled neural radiance fields for
  object categories},'' in \emph{{Proceedings of the IEEE/CVF International
  Conference on Computer Vision}}, 2021, pp. 12\,949--12\,958.

\bibitem{park2021hypernerf}
K.~Park, U.~Sinha, P.~Hedman, J.~T. Barron, S.~Bouaziz, D.~B. Goldman,
  R.~Martin-Brualla, and S.~M. Seitz, ``{HyperNeRF: A higher-dimensional
  representation for topologically varying neural radiance fields},'' \emph{ACM
  Transactions on Graphics (TOG)}, vol.~40, no.~6, pp. 1--12, 2021.

\bibitem{yan2023nerf}
Z.~Yan, C.~Li, and G.~H. Lee, ``{NeRF-DS: Neural Radiance Fields for Dynamic
  Specular Objects},'' in \emph{Proceedings of the IEEE/CVF Conference on
  Computer Vision and Pattern Recognition}, 2023, pp. 8285--8295.

\bibitem{pumarola2021d}
A.~Pumarola, E.~Corona, G.~Pons-Moll, and F.~Moreno-Noguer, ``{D-NeRF: Neural
  radiance fields for dynamic scenes},'' in \emph{{Proceedings of the IEEE/CVF
  Conference on Computer Vision and Pattern Recognition}}, 2021, pp.
  10\,318--10\,327.

\bibitem{park2021nerfies}
K.~Park, U.~Sinha, J.~T. Barron, S.~Bouaziz, D.~B. Goldman, S.~M. Seitz, and
  R.~Martin-Brualla, ``{Nerfies: Deformable neural radiance fields},'' in
  \emph{{Proceedings of the IEEE/CVF International Conference on Computer
  Vision}}, 2021, pp. 5865--5874.

\bibitem{yuan2022neural}
Y.-J. Yuan, Y.-K. Lai, Y.-H. Huang, L.~Kobbelt, and L.~Gao, ``{Neural radiance
  fields from sparse RGB-D images for high-quality view synthesis},''
  \emph{IEEE Transactions on Pattern Analysis and Machine Intelligence}, 2022.

\bibitem{pan2022activenerf}
X.~Pan, Z.~Lai, S.~Song, and G.~Huang, ``{ActiveNeRF: Learning Where to See
  with Uncertainty Estimation},'' in \emph{{Computer Vision--ECCV 2022: 17th
  European Conference, Tel Aviv, Israel, October 23--27, 2022, Proceedings,
  Part XXXIII}}.\hskip 1em plus 0.5em minus 0.4em\relax Springer, 2022, pp.
  230--246.

\bibitem{wang2023sparsenerf}
G.~Wang, Z.~Chen, C.~C. Loy, and Z.~Liu, ``{SparseNeRF: Distilling Depth
  Ranking for Few-shot Novel View Synthesis},'' \emph{arXiv preprint
  arXiv:2303.16196}, 2023.

\bibitem{mildenhall2019local}
B.~Mildenhall, P.~P. Srinivasan, R.~Ortiz-Cayon, N.~K. Kalantari,
  R.~Ramamoorthi, R.~Ng, and A.~Kar, ``{Local light field fusion: Practical
  view synthesis with prescriptive sampling guidelines},'' \emph{ACM
  Transactions on Graphics (TOG)}, vol.~38, no.~4, pp. 1--14, 2019.

\bibitem{jensen2014large}
R.~Jensen, A.~Dahl, G.~Vogiatzis, E.~Tola, and H.~Aan{\ae}s, ``{Large scale
  multi-view stereopsis evaluation},'' in \emph{{Proceedings of the IEEE
  conference on computer vision and pattern recognition}}, 2014, pp. 406--413.

\bibitem{yu2023dylin}
H.~Yu, J.~Julin, Z.~A. Milacski, K.~Niinuma, and L.~A. Jeni, ``{DyLiN: Making
  Light Field Networks Dynamic},'' in \emph{Proceedings of the IEEE/CVF
  Conference on Computer Vision and Pattern Recognition}, 2023, pp.
  12\,397--12\,406.

\bibitem{ramasinghe2023blirf}
S.~Ramasinghe, V.~Shevchenko, G.~Avraham, and A.~V.~D. Hengel, ``{BLiRF:
  Bandlimited Radiance Fields for Dynamic Scene Modeling},'' 2023.

\bibitem{khalid2023refinerf}
S.~Khalid and F.~Rudzicz, ``{RefiNeRF: Modelling dynamic neural radiance fields
  with inconsistent or missing camera parameters},'' 2023.

\bibitem{rao2022icarus}
C.~Rao, H.~Yu, H.~Wan, J.~Zhou, Y.~Zheng, M.~Wu, Y.~Ma, A.~Chen, B.~Yuan,
  P.~Zhou \emph{et~al.}, ``{ICARUS: A Specialized Architecture for Neural
  Radiance Fields Rendering},'' \emph{ACM Transactions on Graphics (TOG)},
  vol.~41, no.~6, pp. 1--14, 2022.

\bibitem{stroud2006boundary}
I.~Stroud, \emph{{Boundary representation modelling techniques}}.\hskip 1em
  plus 0.5em minus 0.4em\relax Springer Science \& Business Media, 2006.

\bibitem{heikkinen2018review}
T.~Heikkinen, J.~Johansson, and F.~Elgh, ``{Review of CAD-model capabilities
  and restrictions for multidisciplinary use},'' \emph{Computer-Aided Design
  and Applications}, vol.~15, no.~4, pp. 509--519, 2018.

\bibitem{louhichi2015cad}
B.~Louhichi, G.~N. Abenhaim, and A.~S. Tahan, ``{CAD/CAE integration: updating
  the CAD model after a FEM analysis},'' \emph{The International Journal of
  Advanced Manufacturing Technology}, vol.~76, pp. 391--400, 2015.

\bibitem{metzer2022latent}
G.~Metzer, E.~Richardson, O.~Patashnik, R.~Giryes, and D.~Cohen-Or,
  ``{Latent-NeRF for Shape-Guided Generation of 3D Shapes and Textures},''
  \emph{arXiv preprint arXiv:2211.07600}, 2022.

\bibitem{poole2022dreamfusion}
B.~Poole, A.~Jain, J.~T. Barron, and B.~Mildenhall, ``{Dreamfusion: Text-to-3d
  using 2d diffusion},'' \emph{arXiv preprint arXiv:2209.14988}, 2022.

\bibitem{saharia2022photorealistic}
C.~Saharia, W.~Chan, S.~Saxena, L.~Li, J.~Whang, E.~L. Denton, K.~Ghasemipour,
  R.~Gontijo~Lopes, B.~Karagol~Ayan, T.~Salimans \emph{et~al.},
  ``{Photorealistic text-to-image diffusion models with deep language
  understanding},'' \emph{Advances in Neural Information Processing Systems},
  vol.~35, pp. 36\,479--36\,494, 2022.

\bibitem{lin2022magic3d}
C.-H. Lin, J.~Gao, L.~Tang, T.~Takikawa, X.~Zeng, X.~Huang, K.~Kreis,
  S.~Fidler, M.-Y. Liu, and T.-Y. Lin, ``{Magic3D: High-Resolution Text-to-3D
  Content Creation},'' \emph{arXiv preprint arXiv:2211.10440}, 2022.

\bibitem{le2023differentiable}
S.~Le~Cleac'h, H.-X. Yu, M.~Guo, T.~Howell, R.~Gao, J.~Wu, Z.~Manchester, and
  M.~Schwager, ``{Differentiable Physics Simulation of Dynamics-Augmented
  Neural Objects},'' \emph{IEEE Robotics and Automation Letters}, vol.~8,
  no.~5, pp. 2780--2787, 2023.

\bibitem{li20223d}
Y.~Li, S.~Li, V.~Sitzmann, P.~Agrawal, and A.~Torralba, ``{3d neural scene
  representations for visuomotor control},'' in \emph{{Conference on Robot
  Learning}}.\hskip 1em plus 0.5em minus 0.4em\relax PMLR, 2022, pp. 112--123.

\bibitem{wang2022clip}
C.~Wang, M.~Chai, M.~He, D.~Chen, and J.~Liao, ``{CLIP-NeRF: Text-and-Image
  Driven Manipulation of Neural Radiance Fields},'' in \emph{{2022 IEEE/CVF
  Conference on Computer Vision and Pattern Recognition (CVPR)}}.\hskip 1em
  plus 0.5em minus 0.4em\relax IEEE, 2022, pp. 3825--3834.

\bibitem{gavish2015evaluating}
N.~Gavish, T.~Guti{\'e}rrez, S.~Webel, J.~Rodr{\'\i}guez, M.~Peveri,
  U.~Bockholt, and F.~Tecchia, ``{Evaluating virtual reality and augmented
  reality training for industrial maintenance and assembly tasks},''
  \emph{Interactive Learning Environments}, vol.~23, no.~6, pp. 778--798, 2015.

\bibitem{fracaro2021towards}
S.~G. Fracaro, P.~Chan, T.~Gallagher, Y.~Tehreem, R.~Toyoda, K.~Bernaerts,
  J.~Glassey, T.~Pfeiffer, B.~Slof, S.~Wachsmuth \emph{et~al.}, ``{Towards
  design guidelines for virtual reality training for the chemical industry},''
  \emph{Education for Chemical Engineers}, vol.~36, pp. 12--23, 2021.

\bibitem{byravan2022nerf2real}
A.~Byravan, J.~Humplik, L.~Hasenclever, A.~Brussee, F.~Nori, T.~Haarnoja,
  B.~Moran, S.~Bohez, F.~Sadeghi, B.~Vujatovic \emph{et~al.}, ``{NeRF2Real:
  Sim2real transfer of vision-guided bipedal motion skills using neural
  radiance fields},'' in \emph{2023 IEEE International Conference on Robotics
  and Automation (ICRA)}.\hskip 1em plus 0.5em minus 0.4em\relax IEEE, 2023,
  pp. 9362--9369.

\bibitem{deng2022fov}
N.~Deng, Z.~He, J.~Ye, B.~Duinkharjav, P.~Chakravarthula, X.~Yang, and Q.~Sun,
  ``{Fov-NeRF: Foveated neural radiance fields for virtual reality},''
  \emph{IEEE Transactions on Visualization and Computer Graphics}, vol.~28,
  no.~11, pp. 3854--3864, 2022.

\bibitem{adamkiewicz2022vision}
M.~Adamkiewicz, T.~Chen, A.~Caccavale, R.~Gardner, P.~Culbertson, J.~Bohg, and
  M.~Schwager, ``{Vision-only robot navigation in a neural radiance world},''
  \emph{IEEE Robotics and Automation Letters}, vol.~7, no.~2, pp. 4606--4613,
  2022.

\bibitem{tancik2022block}
M.~Tancik, V.~Casser, X.~Yan, S.~Pradhan, B.~Mildenhall, P.~P. Srinivasan,
  J.~T. Barron, and H.~Kretzschmar, ``{Block-NeRF: Scalable large scene neural
  view synthesis},'' in \emph{{Proceedings of the IEEE/CVF Conference on
  Computer Vision and Pattern Recognition}}, 2022, pp. 8248--8258.

\bibitem{haque2023instruct}
A.~Haque, M.~Tancik, A.~A. Efros, A.~Holynski, and A.~Kanazawa,
  ``{Instruct-NeRF2NeRF: Editing 3D Scenes with Instructions},'' \emph{arXiv
  preprint arXiv:2303.12789}, 2023.

\bibitem{li2023pac}
X.~Li, Y.-L. Qiao, P.~Y. Chen, K.~M. Jatavallabhula, M.~Lin, C.~Jiang, and
  C.~Gan, ``{PAC-NeRF: Physics Augmented Continuum Neural Radiance Fields for
  Geometry-Agnostic System Identification},'' \emph{arXiv preprint
  arXiv:2303.05512}, 2023.

\bibitem{brena2017evolution}
R.~F. Brena, J.~P. Garc{\'\i}a-V{\'a}zquez, C.~E. Galv{\'a}n-Tejada,
  D.~Mu{\~n}oz-Rodriguez, C.~Vargas-Rosales, and J.~Fangmeyer, ``{Evolution of
  indoor positioning technologies: A survey},'' \emph{Journal of Sensors}, vol.
  2017, 2017.

\bibitem{lin2021barf}
C.-H. Lin, W.-C. Ma, A.~Torralba, and S.~Lucey, ``{Barf: Bundle-adjusting
  neural radiance fields},'' in \emph{{Proceedings of the IEEE/CVF
  International Conference on Computer Vision}}, 2021, pp. 5741--5751.

\bibitem{sun2021neuralrecon}
J.~Sun, Y.~Xie, L.~Chen, X.~Zhou, and H.~Bao, ``{NeuralRecon: Real-time
  coherent 3D reconstruction from monocular video},'' in \emph{{Proceedings of
  the IEEE/CVF Conference on Computer Vision and Pattern Recognition}}, 2021,
  pp. 15\,598--15\,607.

\bibitem{zhu2022nice}
Z.~Zhu, S.~Peng, V.~Larsson, W.~Xu, H.~Bao, Z.~Cui, M.~R. Oswald, and
  M.~Pollefeys, ``{Nice-slam: Neural implicit scalable encoding for slam},'' in
  \emph{{Proceedings of the IEEE/CVF Conference on Computer Vision and Pattern
  Recognition}}, 2022, pp. 12\,786--12\,796.

\bibitem{rosinol2022nerf}
A.~Rosinol, J.~J. Leonard, and L.~Carlone, ``{NeRF-SLAM: Real-Time Dense
  Monocular SLAM with Neural Radiance Fields},'' \emph{arXiv preprint
  arXiv:2210.13641}, 2022.

\bibitem{zhu2022latitude}
Z.~Zhu, Y.~Chen, Z.~Wu, C.~Hou, Y.~Shi, C.~Li, P.~Li, H.~Zhao, and G.~Zhou,
  ``{LATITUDE: Robotic Global Localization with Truncated Dynamic Low-pass
  Filter in City-scale NeRF},'' in \emph{2023 IEEE International Conference on
  Robotics and Automation (ICRA)}.\hskip 1em plus 0.5em minus 0.4em\relax IEEE,
  2023, pp. 8326--8332.

\bibitem{singh2016big}
K.~K. Singh, G.~Chen, J.~B. Vogler, and R.~K. Meentemeyer, ``{When big data are
  too much: Effects of LiDAR returns and point density on estimation of forest
  biomass},'' \emph{IEEE Journal of Selected Topics in Applied Earth
  Observations and Remote Sensing}, vol.~9, no.~7, pp. 3210--3218, 2016.

\bibitem{deibe2018big}
D.~Deibe, M.~Amor, and R.~Doallo, ``{Big data storage technologies: a case
  study for web-based LiDAR visualization},'' in \emph{{2018 IEEE International
  Conference on Big Data (Big Data)}}.\hskip 1em plus 0.5em minus 0.4em\relax
  IEEE, 2018, pp. 3831--3840.

\bibitem{kitchin2016makes}
R.~Kitchin and G.~McArdle, ``{What makes Big Data, Big Data? Exploring the
  ontological characteristics of 26 datasets},'' \emph{Big Data \& Society},
  vol.~3, no.~1, p. 2053951716631130, 2016.

\bibitem{rakotosaona2023nerfmeshing}
M.-J. Rakotosaona, F.~Manhardt, D.~M. Arroyo, M.~Niemeyer, A.~Kundu, and
  F.~Tombari, ``{NeRFMeshing: Distilling Neural Radiance Fields into
  Geometrically-Accurate 3D Meshes},'' \emph{arXiv preprint arXiv:2303.09431},
  2023.

\bibitem{ichnowski2021dex}
J.~Ichnowski, Y.~Avigal, J.~Kerr, and K.~Goldberg, ``{Dex-NeRF: Using a Neural
  Radiance Field to Grasp Transparent Objects},'' in \emph{Conference on Robot
  Learning}.\hskip 1em plus 0.5em minus 0.4em\relax PMLR, 2022, pp. 526--536.

\bibitem{ma2022deblur}
L.~Ma, X.~Li, J.~Liao, Q.~Zhang, X.~Wang, J.~Wang, and P.~V. Sander,
  ``{Deblur-NeRF: Neural radiance fields from blurry images},'' in
  \emph{{Proceedings of the IEEE/CVF Conference on Computer Vision and Pattern
  Recognition}}, 2022, pp. 12\,861--12\,870.

\bibitem{mari2022sat}
R.~Mar{\'\i}, G.~Facciolo, and T.~Ehret, ``{Sat-NeRF: Learning multi-view
  satellite photogrammetry with transient objects and shadow modeling using rpc
  cameras},'' in \emph{{Proceedings of the IEEE/CVF Conference on Computer
  Vision and Pattern Recognition}}, 2022, pp. 1311--1321.

\bibitem{weng2022humannerf}
C.-Y. Weng, B.~Curless, P.~P. Srinivasan, J.~T. Barron, and
  I.~Kemelmacher-Shlizerman, ``{HumanNeRF: Free-viewpoint rendering of moving
  people from monocular video},'' in \emph{{Proceedings of the IEEE/CVF
  Conference on Computer Vision and Pattern Recognition}}, 2022, pp.
  16\,210--16\,220.

\bibitem{peshkin2001cobot}
M.~A. Peshkin, J.~E. Colgate, W.~Wannasuphoprasit, C.~A. Moore, R.~B.
  Gillespie, and P.~Akella, ``{Cobot architecture},'' \emph{IEEE Transactions
  on Robotics and Automation}, vol.~17, no.~4, pp. 377--390, 2001.

\bibitem{gao2022mps}
X.~Gao, J.~Yang, J.~Kim, S.~Peng, Z.~Liu, and X.~Tong, ``{MPS-NeRF:
  Generalizable 3D Human Rendering from Multiview Images},'' \emph{IEEE
  Transactions on Pattern Analysis and Machine Intelligence}, 2022.

\bibitem{neumann2019aerial}
P.~P. Neumann, H.~Kohlhoff, D.~H{\"u}llmann, D.~Krentel, M.~Kluge,
  M.~Dzierli{\'n}ski, A.~J. Lilienthal, and M.~Bartholmai, ``{Aerial-based gas
  tomography--from single beams to complex gas distributions},'' \emph{European
  Journal of Remote Sensing}, vol.~52, no. sup3, pp. 2--16, 2019.

\bibitem{zang2021intratomo}
G.~Zang, R.~Idoughi, R.~Li, P.~Wonka, and W.~Heidrich, ``{IntraTomo:
  self-supervised learning-based tomography via sinogram synthesis and
  prediction},'' in \emph{{Proceedings of the IEEE/CVF International Conference
  on Computer Vision}}, 2021, pp. 1960--1970.

\bibitem{yen2022nerf}
L.~Yen-Chen, P.~Florence, J.~T. Barron, T.-Y. Lin, A.~Rodriguez, and P.~Isola,
  ``{NeRF-Supervision: Learning dense object descriptors from neural radiance
  fields},'' in \emph{{2022 International Conference on Robotics and Automation
  (ICRA)}}.\hskip 1em plus 0.5em minus 0.4em\relax IEEE, 2022, pp. 6496--6503.

\bibitem{truong2020glu}
P.~Truong, M.~Danelljan, and R.~Timofte, ``{GLU-Net: Global-local universal
  network for dense flow and correspondences},'' in \emph{{Proceedings of the
  IEEE/CVF conference on computer vision and pattern recognition}}, 2020, pp.
  6258--6268.

\bibitem{truong2020gocor}
P.~Truong, M.~Danelljan, L.~V. Gool, and R.~Timofte, ``{GOCor: Bringing
  globally optimized correspondence volumes into your neural network},''
  \emph{Advances in Neural Information Processing Systems}, vol.~33, pp.
  14\,278--14\,290, 2020.

\bibitem{truong2021learning}
P.~Truong, M.~Danelljan, L.~Van~Gool, and R.~Timofte, ``{Learning accurate
  dense correspondences and when to trust them},'' in \emph{{Proceedings of the
  IEEE/CVF Conference on Computer Vision and Pattern Recognition}}, 2021, pp.
  5714--5724.

\bibitem{goli2022nerf2nerf}
L.~Goli, D.~Rebain, S.~Sabour, A.~Garg, and A.~Tagliasacchi, ``{nerf2nerf:
  Pairwise Registration of Neural Radiance Fields},'' in \emph{2023 IEEE
  International Conference on Robotics and Automation (ICRA)}.\hskip 1em plus
  0.5em minus 0.4em\relax IEEE, 2023, pp. 9354--9361.

\bibitem{zhou2023nerf}
A.~Zhou, M.~J. Kim, L.~Wang, P.~Florence, and C.~Finn, ``{NeRF in the Palm of
  Your Hand: Corrective Augmentation for Robotics via Novel-View Synthesis},''
  in \emph{Proceedings of the IEEE/CVF Conference on Computer Vision and
  Pattern Recognition}, 2023, pp. 17\,907--17\,917.

\bibitem{ge2022neural}
Y.~Ge, H.~Behl, J.~Xu, S.~Gunasekar, N.~Joshi, Y.~Song, X.~Wang, L.~Itti, and
  V.~Vineet, ``{Neural-Sim: Learning to Generate Training Data with NeRF},'' in
  \emph{{Computer Vision--ECCV 2022: 17th European Conference, Tel Aviv,
  Israel, October 23--27, 2022, Proceedings, Part XXIII}}.\hskip 1em plus 0.5em
  minus 0.4em\relax Springer, 2022, pp. 477--493.

\bibitem{xu2023}
X.~Xu, Y.~Yang, K.~Mo, B.~Pan, L.~Yi, and L.~Guibas, ``{JacobiNeRF: NeRF
  shaping with mutual information gradients},'' in \emph{{Proceedings of the
  IEEE/CVF Conference on Computer Vision and Pattern Recognition}}, 2023, pp.
  16\,498--16\,507.

\bibitem{muller2022instant}
T.~M{\"u}ller, A.~Evans, C.~Schied, and A.~Keller, ``{Instant neural graphics
  primitives with a multiresolution hash encoding},'' \emph{ACM Transactions on
  Graphics (ToG)}, vol.~41, no.~4, pp. 1--15, 2022.

\bibitem{DBLP:journals/corr/abs-2111-11426}
\BIBentryALTinterwordspacing
Y.~Xie, T.~Takikawa, S.~Saito, O.~Litany, S.~Yan, N.~Khan, F.~Tombari,
  J.~Tompkin, V.~Sitzmann, and S.~Sridhar, ``Neural fields in visual computing
  and beyond,'' \emph{CoRR}, vol. abs/2111.11426, 2021. [Online]. Available:
  \url{https://arxiv.org/abs/2111.11426}
\BIBentrySTDinterwordspacing

\bibitem{M_ller_2022}
\BIBentryALTinterwordspacing
T.~Müller, A.~Evans, C.~Schied, and A.~Keller, ``{Instant neural graphics
  primitives with a multiresolution hash encoding},'' \emph{{ACM} Transactions
  on Graphics}, vol.~41, no.~4, pp. 1--15, jul 2022. [Online]. Available:
  \url{https://doi.org/10.1145%2F3528223.3530127}
\BIBentrySTDinterwordspacing

\bibitem{hore2010image}
A.~Hore and D.~Ziou, ``{Image quality metrics: PSNR vs. SSIM},'' in \emph{{2010
  20th international conference on pattern recognition}}.\hskip 1em plus 0.5em
  minus 0.4em\relax IEEE, 2010, pp. 2366--2369.

\bibitem{fortun2015}
D.~Fortun, P.~Bouthemy, and C.~Kervrann, ``{Optical flow modeling and
  computation: A survey},'' \emph{Computer Vision and Image Understanding},
  vol. 134, pp. 1--21, 2015.

\bibitem{stockhammer2003h}
T.~Stockhammer, M.~M. Hannuksela, and T.~Wiegand, ``{H. 264/AVC in wireless
  environments},'' \emph{IEEE Transactions on Circuits and Systems for Video
  technology}, vol.~13, no.~7, pp. 657--673, 2003.

\bibitem{heindl20193d}
C.~Heindl, S.~Zambal, T.~Ponitz, A.~Pichler, and J.~Scharinger, ``{3D robot
  pose estimation from 2D images},'' \emph{arXiv preprint arXiv:1902.04987},
  2019.

\bibitem{fan2019brief}
L.~Fan, F.~Zhang, H.~Fan, and C.~Zhang, ``Brief review of image denoising
  techniques,'' \emph{Visual Computing for Industry, Biomedicine, and Art},
  vol.~2, pp. 1--12, 2019.

\bibitem{ilesanmi2021methods}
A.~E. Ilesanmi and T.~O. Ilesanmi, ``Methods for image denoising using
  convolutional neural network: a review,'' \emph{Complex \& Intelligent
  Systems}, vol.~7, no.~5, pp. 2179--2198, 2021.

\bibitem{bindal2022systematic}
N.~Bindal, R.~S. Ghumaan, P.~J.~S. Sohi, N.~Sharma, H.~Joshi, and B.~Garg, ``A
  systematic review of state-of-the-art noise removal techniques in digital
  images,'' \emph{Multimedia Tools and Applications}, vol.~81, no.~22, pp.
  31\,529--31\,552, 2022.

\bibitem{jamil2008noise}
N.~Jamil, T.~M.~T. Sembok, and Z.~A. Bakar, ``Noise removal and enhancement of
  binary images using morphological operations,'' in \emph{2008 International
  Symposium on Information Technology}, vol.~4.\hskip 1em plus 0.5em minus
  0.4em\relax IEEE, 2008, pp. 1--6.

\bibitem{kerr2023lerf}
J.~Kerr, C.~M. Kim, K.~Goldberg, A.~Kanazawa, and M.~Tancik, ``{LERF: Language
  Embedded Radiance Fields},'' \emph{arXiv preprint arXiv:2303.09553}, 2023.

\end{thebibliography}

\begin{IEEEbiography}[{\includegraphics[width=1in,height=1.25in,clip,keepaspectratio]{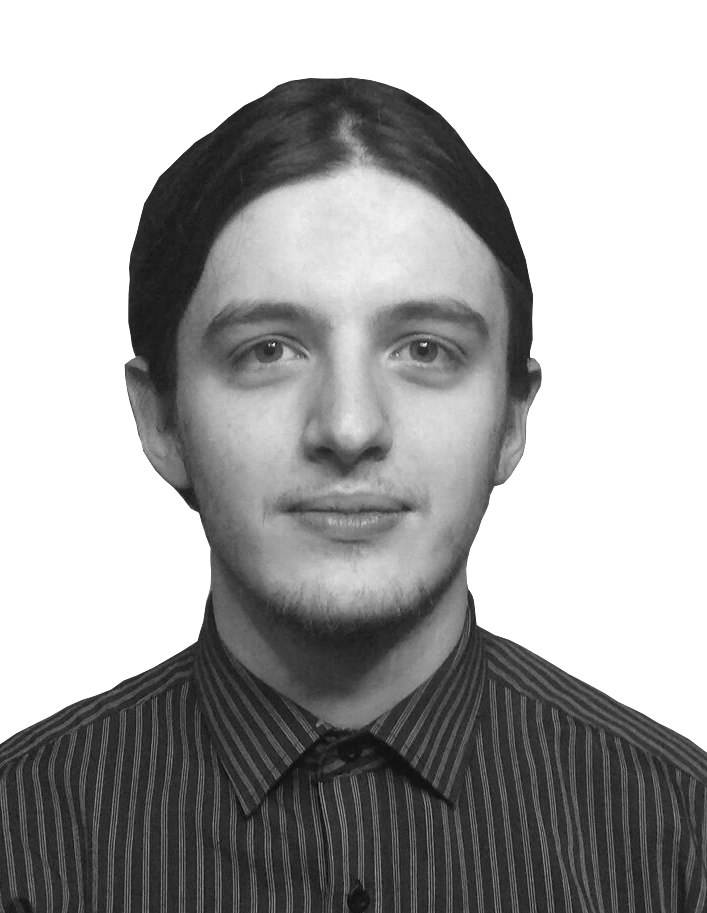}}]{Eugen \v{S}lapak} is an assistant professor at the Technical University of Ko\v{s}ice, Slovakia. His PhD thesis focused on 5G HetNet physical topology design using advanced machine learning algorithms. He was a guest researcher at King`s College London under supervision of Prof. Mischa Dohler. Currently, he works with the research team at the university's Intelligent Information Systems Laboratory (http://iislab.kpi.fei.tuke.sk/), and his research interests include radio access network simulation, computer vision, metaheuristic optimization and machine learning. He serves as the regular reviewer in several recognized IEEE Transactions journals.
\end{IEEEbiography}

\begin{IEEEbiography}[{\includegraphics[width=1in,height=1.25in,clip,keepaspectratio]{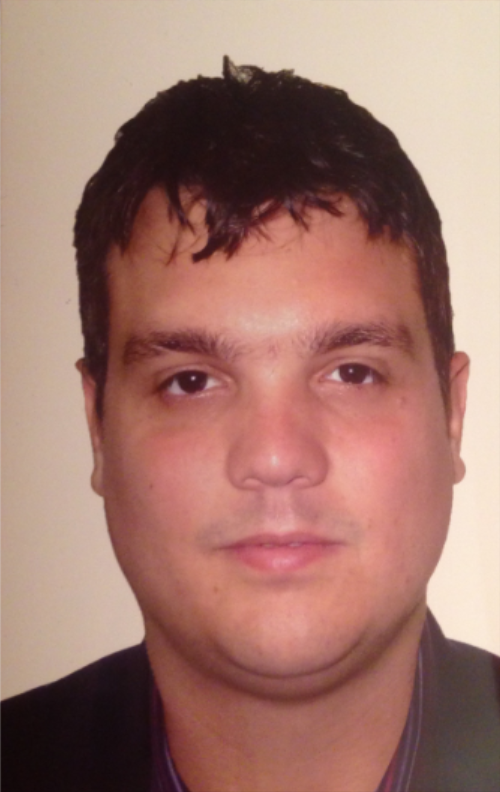}}]{Enric Pardo} is a 5G research and technology associate at LIST. Prior to that, he received a PhD at the Centre for Telecommunications Research at King's College London, U.K. in the area of multiconnectivity in 5G networks. He received his B.Sc. degree in telecommunications engineering from Universitat Politecnica de Catalunya (Spain). His research interests include applied mathematics in 5G communications for network performance evaluation and the coexistence of terrestrial and aerial users. 
\end{IEEEbiography}

\begin{IEEEbiography}[{\includegraphics[width=1in,height=1.25in,clip,keepaspectratio]{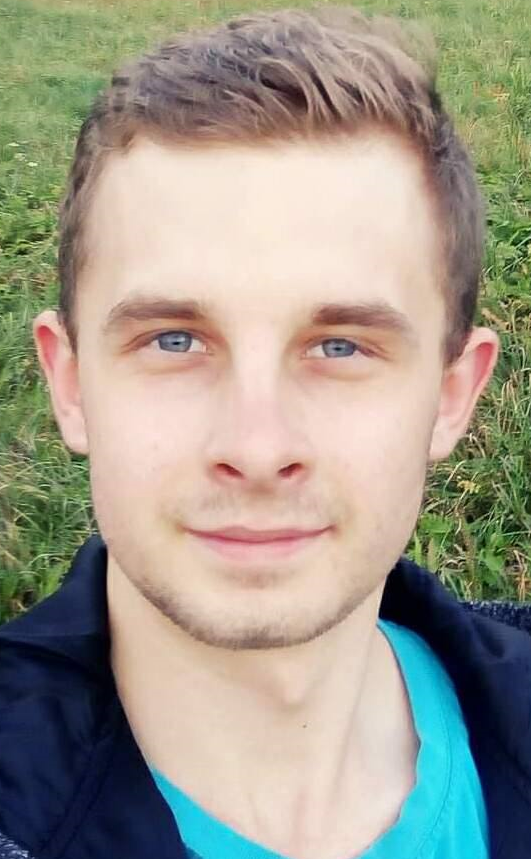}}]{ Matúš Dopiriak} 
 received his BSc. and MSc. degrees in informatics from the Technical University of Ko\v{s}ice, Slovakia. He completed the studies at the Department of Computers and Informatics in the years 2020 and 2022, respectively. His diploma thesis centered on the implementation of CNN-based architectures to monitor traffic from the perspective of an UAV. Currently, he is a doctoral candidate under the supervision of Prof. Ing. Juraj Gazda, PhD. His present research endeavors revolve around the integration of neural architectures, with a specific emphasis on the application of neural radiance fields, in robotics and autonomous mobility of vehicles, particularly in the context of edge computing.
\end{IEEEbiography}

\vfill

\begin{IEEEbiography}[{\includegraphics[width=1in,height=1.25in,clip,keepaspectratio]{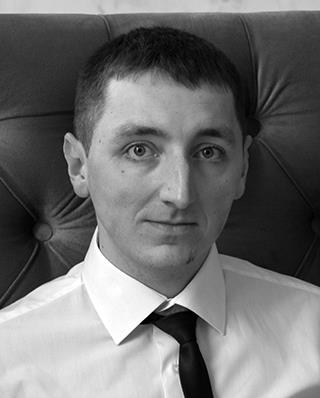}}]{Taras Maksymyuk} (Member, IEEE) received a PhD degree in telecommunication systems and networks in 2015 from Lviv Polytechnic National University. He is currently a research professor with the Telecommunications Department, Lviv Polytechnic National University, Lviv, Ukraine. He performed his postdoctoral fellowship at the Internet of Things and Artificial Intelligence Laboratory, Korea University. He is currently an associate editor of \emph{IEEE Communications Magazine} and an editor of \emph{Wireless Communications and Mobile Computing}. His research interests include 5G heterogeneous networks, software-defined networks, the Internet of Things, blockchain, big data and artificial intelligence.
\end{IEEEbiography}

\begin{IEEEbiography}[{\includegraphics[width=1in,height=1.25in,clip,keepaspectratio]{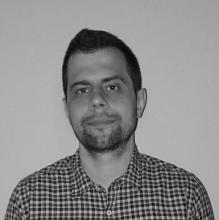}}]{Juraj Gazda} is currently a professor with the Faculty of Electrical Engineering and Informatics at the Technical University of Ko\v{s}ice (TUKE), Slovakia. He has been a guest researcher at Ramon Llull University, Barcelona, and the Technical University of Hamburg-Harburg. He has been involved in the development for Nokia Siemens Networks (NSN) and Ericsson. In 2017, he was recognized as the Best Young Scientist at TUKE. Currently, he serves as the editor of \emph{KSII Transactions on Internet and Information Systems} and as a guest editor of \emph{Wireless Communications and Mobile Computing (Wiley)}. His research interests include techno-economic aspects of 5G/6G networks, computer vision, and artificial intelligence.
\end{IEEEbiography}

\vfill

\end{document}